\documentclass[runningheads]{llncs}
\newif\ifdraft

\newif\ifrisn
\risntrue

\usepackage{a4}
\usepackage{amssymb}
\usepackage{amsmath}
\usepackage[lined,boxed,linesnumbered]{algorithm2e}
\usepackage{array}
\usepackage{dcolumn}
\usepackage[acronym]{glossaries}
\usepackage{graphicx}
\usepackage{hyperref}
\usepackage{imakeidx}  
\usepackage{latexsym}
\usepackage{mfirstuc}
\usepackage{multicol}
\usepackage{multirow}
\usepackage{rotating}
\usepackage{soul}
\usepackage{tabularx,booktabs}
\usepackage{url}

\usepackage[utf8]{inputenc}
\usepackage{outlines}
\usepackage{wasysym}

\usepackage{etoolbox}
\usepackage[table]{xcolor}

\newcommand{\blue}[1]{\ifdraft{\leavevmode\color{blue}{#1}}\else{\leavevmode\color{black}{#1}}\fi}

\newcommand{\replace}[2]{\ifdraft{\st{#1} \leavevmode\color{blue}{#2}}\else{\textcolor{black}{#2}}\fi}
\newcommand{\removeifspaceneeded}[1]{\ifrisn{\vspace{0ex}}\else{#1}\fi}

\newcommand{\FP}{\mathrm{FP}}
\newcommand{\FN}{\mathrm{FN}}

\newcommand{\TPR}{\mathrm{TPR}}
\newcommand{\FPR}{\mathrm{FPR}}

\DeclareMathOperator{\aae}{AE}

\DeclareMathOperator{\rae}{RAE}

\newcolumntype{.}{U{.}{.}{-1}}
\newcolumntype{M}[1]{>{\centering\arraybackslash}m{#1}}
\newcolumntype{N}{@{}m{0pt}@{}}
\newcolumntype{C}[1]{>{\centering\let\newline\\\arraybackslash\hspace{0pt}}m{#1}}
\newcolumntype{Y}{>{\centering\arraybackslash}X}

\newacronym{BQ}{BQ}{Binary quantification}
\newacronym{SLQ}{SLQ}{Single-label quantification}
\newacronym{MLQ}{MLQ}{Multi-label quantification}

\begin{document}
\title{Re-Assessing the ``Classify and Count'' Quantification Method}


\author{Alejandro Moreo\orcidID{0000-0002-0377-1025} \and Fabrizio
Sebastiani\orcidID{0000-0003-4221-6427} }
\authorrunning{Alejandro Moreo and Fabrizio Sebastiani}
%
\institute{Istituto di Scienza e Tecnologie dell'Informazione \\
Consiglio Nazionale delle Ricerche \\
56124 Pisa, Italy \\
\email{firstname.lastname@isti.cnr.it} }

\maketitle 
\begin{abstract}
  \emph{Learning to quantify} (a.k.a.\ \emph{quantification}) is a
  task concerned with training unbiased estimators of class prevalence
  via supervised learning. This task originated with the observation
  that ``Classify and Count'' (CC), the trivial method of obtaining
  class prevalence estimates, is often a biased estimator, and thus
  delivers suboptimal quantification accuracy. Following this
  observation, several methods for learning to quantify have been
  proposed and have been shown to outperform CC. In this work we
  contend that previous works have failed to use properly optimised
  versions of CC. We thus reassess the real merits of CC and its
  variants, and argue that, while still inferior to some cutting-edge
  methods, they deliver near-state-of-the-art accuracy once (a)
  hyperparameter optimisation is performed, and (b) this optimisation
  is performed by using a \replace{true quantification loss instead of
  a standard classification-based loss}{truly quantification-oriented
  evaluation protocol}. Experiments on three publicly available binary
  sentiment classification datasets support these conclusions.
  
\end{abstract}


\section{Introduction}
\label{sec:intro}

\noindent \emph{Learning to quantify} (a.k.a.\ \emph{quantification})
consists of training a predictor that returns estimates of the
relative frequency (a.k.a.\ \emph{prevalence}, or \emph{prior
probability}) of the classes of interest in a set of unlabelled data
items, where the predictor has been trained on a set of labelled data
items~\cite{Gonzalez:2017it}.  When applied to text, quantification is
important for several applications, e.g., gauging the collective
satisfaction for a certain product from textual
comments~\cite{Esuli:2020rc}, establishing the popularity of a given
political candidate from blog posts~\cite{Hopkins:2010fk}, predicting
the amount of consensus for a given governmental policy from
tweets~\cite{Borge-Holthoefer:2015dz}, or predicting the amount of
readers who will find a product review helpful~\cite{Card:2018pb}.

The rationale of this task is that many real-life applications of
classification suffer from \emph{distribution
shift}~\cite{Moreno-Torres:2012ay}, the phenomenon according to which
the distribution $p_{y}(U)$ of the labels in the set of unlabelled
test documents $U$ is different from the distribution $p_{y}(L)$ that
the labels have in the set of labelled training documents $L$. It has
been shown that, in the presence of distribution shift, the trivial
strategy of using a standard classifier to classify all the unlabelled
documents in $U$ and counting the documents that have been assigned to
each class (the ``Classify and Count'' (CC) method), delivers poor
class prevalence estimates. The reason is that most supervised
learning methods are based on the IID assumption, which implies
\removeifspaceneeded{(among others)}that the distribution of the
labels is the same in $L$ and $U$. ``Classify and Count'' is
considered a \emph{biased estimator} of class prevalence, since the
goal of standard classifiers is to minimise (assuming for simplicity a
binary setting) \emph{classification} error measures such as
$(\FP+\FN)$, while the goal of a quantifier is to minimise
\emph{quantification} error measures such as $|\FP-\FN|$.  (In this
paper we tackle binary quantification, so $\FP$ and $\FN$ denote the
numbers of false positives and false negatives, resp., from a binary
contingency table.)  Following this observation, several
quantification methods have been proposed, and have been
experimentally shown to outperform CC.
  
In this paper we contend that previous works, when testing advanced
quantification methods, have used as baselines versions of CC that had
not been properly optimised. This means that published results on the
relative merits of CC and other supposedly more advanced methods are
still unreliable.  We thus reassess the real merits of CC by running
extensive experiments (on three publicly available sentiment
classification datasets) in which we compare properly optimised
versions of CC and its three main variants (PCC, ACC, PACC) with a
number of more advanced quantification methods. In these experiments
we properly optimise all quantification methods, i.e., (a) we optimise
their hyperparameters, and (b) we conduct this optimisation \blue{via
a truly quantification-oriented evaluation protocol, which also
involves} minimising a quantification loss rather than a
classification loss.  Our results indicate that, while still inferior
to some cutting-edge quantification methods, CC and its variants
deliver near-state-of-the-art quantification accuracy once
hyperparameter optimisation is performed properly. We make available
all the code and the datasets that we have used for our
experiments.\footnote{\url{https://github.com/AlexMoreo/CC}}
\removeifspaceneeded{The rest of this paper is structured as
follows. In Section~\ref{sec:classifyandcount} we briefly present
Classify and Count and its three main variants. In
Section~\ref{sec:paroptquant} we discuss how proper optimisation of
quantification methods should be performed, and show that these
standards have seldom been adhered to in past quantification
literature when using CC as a baseline.  Section~\ref{sec:experiments}
discusses the new experiments we have run in order to compare CC with
other methods under more strict experimental
standards. Section~\ref{sec:results} discusses the results and the
conclusions that they allow to draw.}


\section{``Classify and Count'' and its variants}
\label{sec:classifyandcount}

\noindent In this paper we use the following notation.  We assume a
binary setting, with the two classes $\mathcal{Y}=\{\oplus,\ominus\}$
standing for \textsf{Positive} and \textsf{Negative}. By $\mathbf{x}$
we denote a document drawn from a domain $\mathcal{X}$ of documents;
by $L\subset \mathcal{X}$ we denote a set of \underline{l}abelled
documents, that we typically use as a training set, while by $U$ we
denote a sample of \underline{u}nlabelled documents, that we typically
use as the sample to quantify on.  By $p_{y}(\sigma)$ we indicate the
true prevalence of class $y$ in sample $\sigma$, by
$\hat{p}_{y}(\sigma)$ we indicate an estimate of this
prevalence\footnote{Consistently with most mathematical literature, we
use the caret symbol (\^\/\/) to indicate estimation.}, and by
$\hat{p}_{y}^{M}(\sigma)$ we indicate the estimate of this prevalence
as obtained via quantification method $M$. Of course, for any method
$M$ it holds that
$\hat{p}_{\ominus}^{\mathit{M}}(U)=(1-\hat{p}_{\oplus}^{\mathit{M}}(U))$.

An obvious way to solve quantification is by aggregating the scores
assigned by a classifier to the unlabelled documents.
We first define two different aggregation methods, one that uses a
``hard'' classifier (i.e., a classifier
$h_{\oplus}:\mathcal{X} \rightarrow \{0,1\}$ that returns binary
decisions, 0 for $\ominus$ and 1 for $\oplus$) and one that uses a
``soft'' classifier (i.e., a classifier
$s_{\oplus}:\mathcal{X} \rightarrow [0,1]$ that returns posterior
probabilities $\Pr(\oplus|\mathbf{x})$, representing the probability
that the classifier attributes to the fact that $\mathbf{x}$ belongs
to the $\oplus$ class). Of course,
$\Pr(\ominus|\mathbf{x})=(1-\Pr(\oplus|\mathbf{x}))$.
The \emph{classify and count} ($\mathrm{CC}$) and the
\emph{probabilistic classify and count} ($\mathrm{PCC}$)
\cite{Bella:2010kx} methods then consist of computing
\begin{align}
  \label{eq:cc}
  \hat{p}_{\oplus}^{\mathrm{CC}}(U)  = \frac{\sum_{\mathbf{x}\in U}
  h_{\oplus}(\mathbf{x})}{|U|} \hspace{5em}
  \hat{p}_{\oplus}^{\mathrm{PCC}}(U)  = \frac{\sum_{\mathbf{x}\in U}
  s_{\oplus}(\mathbf{x})}{|U|}
\end{align}
\noindent Two popular, alternative quantification methods consist of
applying an \emph{adjustment} to the
$\hat{p}_{\oplus}^{\mathrm{CC}}(U)$ and
$\hat{p}_{\oplus}^{\mathrm{PCC}}(U)$ estimates. It is easy to show
that, in the binary case, the true prevalence $p_{\oplus}(U)$ is such
that
\begin{equation}
  \label{eq:exactacc} 
  p_{\oplus}(U) = \frac{\hat{p}_{\oplus}^{\mathrm{CC}}(U) 
  - \FPR_{h}}{\TPR_{h} - \FPR_{h}} 
  \hspace{5em}
  p_{\oplus}(U) = \frac{\hat{p}_{\oplus}^{\mathrm{PCC}}(U) 
  - \FPR_{s}}{\TPR_{s} - \FPR_{s}} 
\end{equation}
\noindent where $\TPR_{h}$ and $\FPR_{h}$ (resp., $\TPR_{s}$ and
$\FPR_{s}$) here stand for the \emph{true positive rate} and
\emph{false positive rate} that the classifier $h_{\oplus}$ (resp.,
$s_{\oplus}$) has on $U$.  The values of $\TPR_{h}$ and $\FPR_{h}$
(resp., $\TPR_{s}$ and $\FPR_{s}$) are unknown, but can be estimated
via $k$-fold cross-validation on the training data.  In the binary
case this amounts to using the results that $h_{\oplus}(\mathbf{x})$
(resp., $s_{\oplus}(\mathbf{x})$) obtains in the $k$-fold
cross-validation (i.e., when $\mathbf{x}$ ranges on the training
documents) in equations
\begin{align}
  \begin{split}
    \label{eq:tprandfpr}
    \hat{\TPR_{h}} = \frac{\sum_{\mathbf{x}\in
    \oplus}h_{\oplus}(\mathbf{x})}{|\oplus|} \hspace{3em}
    \hat{\FPR_{h}} =\frac{\sum_{\mathbf{x}\in
    \ominus}h_{\oplus}(\mathbf{x})}{|\ominus|} \\
    \hat{\TPR_{s}} = \frac{\sum_{\mathbf{x}\in
    \oplus}s_{\oplus}(\mathbf{x})}{|\oplus|} \hspace{3em}
    \hat{\FPR_{s}} =\frac{\sum_{\mathbf{x}\in
    \ominus}s_{\oplus}(\mathbf{x})}{|\ominus|}
  \end{split}
\end{align}
\noindent We obtain $\hat{p}_{\oplus}^{\mathrm{ACC}}(U)$ and
$\hat{p}_{\oplus}^{\mathrm{PACC}}(U)$ estimates, which define the
\emph{adjusted classify and count}
($\mathrm{ACC}$)~\cite{Forman:2008kx} and \emph{probabilistic adjusted
classify and count} ($\mathrm{PACC}$)~\cite{Bella:2010kx}
quantification methods, resp., by replacing $\TPR_{h}$ and $\FPR_{h}$
(resp., $\TPR_{s}$ and $\FPR_{s}$) in Equation~\ref{eq:exactacc} with
their estimates from Equation~\ref{eq:tprandfpr}.


\section{Quantification and parameter optimisation}
\label{sec:paroptquant}


\subsection{Unsuitable parameter optimisation and weak baselines}
\label{sec:weakbaselines}

\noindent The reason why we here reassess CC and its variants we have
described above, is that we believe that, in previous papers where
these methods have been used as baselines, their full potential has
not been realised because of \emph{missing or unsuitable optimisation}
of the hyperparameters of the classifier on which the method is based.

Specifically, both CC and its
variants rely on the output of a previously trained classifier, and
this output usually depends on some hyperparameters. Not only the
quality of this output heavily depends on whether these
hyperparameters have been optimised or not (on some held-out data or
via $k$-fold cross-validation), but \emph{it also depends on what
evaluation measure this optimisation has used as a criterion for model
selection}. In other words, given that hyperparameter optimisation
chooses the value of the parameter that minimises error, it would make
sense that, for a classifier to be used for quantification purposes,
``error'' is measured via a function that evaluates
\emph{quantification} error, and not classification
error. Unfortunately, in most previous quantification papers,
researchers either do not specify whether hyperparameter optimisation
was performed at
all~\cite{Esuli:2014uq,Forman:2008kx,Gonzalez:2016xy,Gonzalez-Castro:2013fk,Hopkins:2010fk,Levin:2017dq,Perez-Gallego:2017wt,Saerens:2002uq},
or leave the hyperparameters at their default
values~\cite{Barranquero:2015fr,Bella:2010kx,Esuli:2015gh,Hassan:2020jb,Milli:2013fk},
or do not specify which evaluation measure they use in hyperparameter
optimisation~\cite{Esuli:2020rc,Gao:2016uq}, or use, for this
optimisation, a classification-based
loss~\cite{Barranquero:2013fk,Perez-Gallego:2019vl}. In retrospect, we
too plead guilty, since some of the papers quoted here are our own.

All this means that CC and their variants, when used as baselines,
have been turned into \emph{weak} baselines, and this
means that the merits of more modern methods relative to them have
possibly been exaggerated, and are thus yet to be assessed
reliably. In this paper we thus engage in a reproducibility study, and
present results from text quantification experiments in which,
contrary to the situations described in the paragraph above, we
compare \emph{carefully optimised} versions of CC and its variants
with a number of (\emph{carefully optimised} versions of) more modern
quantification methods, in an attempt to assess the relative value of
each in a robust way.


\subsection{Quantification-oriented parameter optimisation}
\label{sec:paropt}

\noindent In order to perform quantification-oriented parameter
optimisation we need to be aware that there may exist two types of
parameters that require estimation and/or optimisation, i.e., (a) the
hyperparameters of the classifier on which the quantification method
is based, and (b) the parameters of the quantification method itself.

The way we perform hyperparameter optimisation is the following. We
assume that the dataset comes with a predefined split between a
training set $L$ and a test set $U$. (This assumption is indeed
verified for the datasets we will use in
Section~\ref{sec:experiments}.)  We first partition $L$ into a part
$L_{\mathrm{Tr}}$ that will be used for training purposes and a part
$L_{\mathrm{Va}}$ that will be used as a held-out validation set for
optimising the hyperparameters of the quantifier.
We then extract, from the validation set $L_{\mathrm{Va}}$, several
random validation samples, each characterised by a predefined
prevalence of the $\oplus$ class; here, our goal is allowing the
validation to be conducted on a variety of scenarios characterised by
widely different values of class prevalence, and, as a consequence, by
widely different amounts of distribution shift.\footnote{Note that
this is similar to what we do, say, in classification, where the
different hyperparameter values are tested on many validation
documents; here we test these hyperparameter values on many validation
\emph{samples}, since the objects of study of text quantification are
document samples inasmuch as the objects of study of text
classification are individual documents.} In order to do this, we
extract each validation sample $\sigma$ by randomly undersampling one
or both classes in $L_{\mathrm{Va}}$, in order to obtain a sample with
prespecified class prevalence values. We draw samples with a desired
prevalence value and a fixed amount $q$ of documents; in order to
achieve this, in some cases only one class needs to be undersampled
while in some other cases this needs to happen for both classes. We
use random sampling without replacement if the number of available
examples of $\oplus$ (resp.\ $\ominus$) is greater or equal to the
number of required ones, and with replacement otherwise.
We extract samples with a prevalence of the $\oplus$ class in the set
$\{\pi_{1}, ..., \pi_{n}\}$;
for each of these $n$ values we generate $m$ random samples consisting
of $q$ validation documents each.  Let $\Theta$ be the set of
hyperparameters that we are going to optimise. Given the established
grid of value combinations $\theta_{1}, ..., \theta_{n}$ that we are
going to test for $\Theta$, for each $\theta_{i}$ we do the following,
depending on whether the quantification method has its own parameters
(Case~\ref{item:pars} below) or not
(Case~\ref{item:nopars} below):
\begin{enumerate}

\item \label{item:pars}
  If the quantification method $M$ we are going to optimise requires
  some parameters $\lambda_i$ to be estimated, we first split
  $L_{\mathrm{Tr}}$ into a part $L_{\mathrm{Tr}}^{\mathrm{Tr}}$ and a
  part $L_{\mathrm{Tr}}^{\mathrm{Va}}$, training the classifier on
  $L_{\mathrm{Tr}}^{\mathrm{Tr}}$ using the chosen learner
  parameterised with $\theta_{i}$, and estimate parameters $\lambda_i$
  on $L_{\mathrm{Tr}}^{\mathrm{Va}}$.\footnote{Note that we do
  \emph{not} retrain the classifier on the entire
  $L_{\mathrm{Tr}}$. While this might seem beneficial, since
  $L_{\mathrm{Tr}}$ contains more training data than
  $L_{\mathrm{Tr}}^{\mathrm{Tr}}$, we need to consider that the
  estimates $\hat{\TPR}_{h}$ and $\hat{\FPR}_{h}$ have been computed
  on $L_{\mathrm{Tr}}$ and not on $L_{\mathrm{Tr}}^{\mathrm{Tr}}$.}
  Among the variants of CC, this applies to methods ACC and PACC,
  which require the estimation of (the hard or soft version of) $\TPR$
  and $\FPR$. 
  Other methods used in the experiments of
  Section~\ref{sec:experiments} and that also require some parameter
  to be estimated are HDy and QuaNet (see Section~\ref{sec:advanced}).

\item \label{item:nopars} If the quantification method $M$ we are
  going to optimise does not have any parameter that requires
  estimation, then we train our classifier on $L_{\mathrm{Tr}}$, using
  the chosen learner parameterised with $\theta_{i}$, and use
  quantification method $M$ on all the samples extracted from
  $L_{\mathrm{Va}}$.
  
\end{enumerate}
\noindent In both cases, we measure the quantification error via an
evaluation measure for quantification that combines (e.g., averages)
the results across all the validation samples. As our final value
combination for hyperparameter set $\Theta$ we choose the $\theta_{i}$
for which quantification error is minimum.

Note that, in the above discussion, each time we split a labelled set
into a training set and a validation set for parameter estimation /
optimisation purposes, we could instead perform a $k$-fold
cross-validation; the parameter estimation/optimisation would be more
robust, but the computational cost of the entire process would be $k$
times higher. While the latter method is also, from a methodological
standpoint, an option, in this paper we stick to the former method,
since the entire parameter optimisation process is, from a
computational point of view, already very expensive.


\section{Experiments}
\label{sec:experiments}

%

\noindent In order to conduct our experiments we use the same datasets
and experimental protocol as used
in~\cite{Esuli:2018rm}. Specifically, we run our experiments on three
sentiment classification datasets, i.e., (i) \textsc{IMDB}, the
popular \emph{Large Movie Review Dataset} \cite{Maas2011}; (ii)
\textsc{Kindle}, a set of reviews of Kindle e-book
readers~\cite{Esuli:2018rm}, and (iii) \textsc{HP}, a set of reviews
of the books from the Harry Potter
series~\cite{Esuli:2018rm}.\footnote{The three datasets are available
at \url{https://doi.org/10.5281/zenodo.4117827} in pre-processed form.
The raw versions of the \textsc{HP} and \textsc{Kindle} datasets can
be accessed from \url{http://hlt.isti.cnr.it/quantification/}, while
the raw version of \textsc{IMDB} can be found at
\url{https://ai.stanford.edu/~amaas/data/sentiment/}.} For all
datasets we adopt the same split between training set $L$ and test set
$U$ as in~\cite{Esuli:2018rm}. The \textsc{IMDB}, \textsc{Kindle}, and
\textsc{HP} datasets are examples of balanced, imbalanced, and
severely imbalanced datasets, since the prevalence values of the
$\oplus$ class in the training set $L$ are 0.500, 0.917, 0.982,
resp. Some basic statistics from these datasets are reported in
Table~\ref{tab:datasets}. We refer the reader to~\cite{Esuli:2018rm}
for more details on the genesis of these datasets.

\begin{table}[t]
  \caption{The three datasets used in our experiments; the columns
  indicate the class prevalence values of the $\oplus$ and $\ominus$
  classes, and the numbers of documents contained in the training set
  $L$ and the test set $U$.}
  \begin{center}
    {\scriptsize
    \begin{tabular}{|r||c|c|c|c|c|c|} \hline & $\oplus$ & $\ominus$ &
      $L$ & $L_{\mathrm{Tr}}$ &
                                $L_{\mathrm{Va}}$ & $U$ \\
      \hline \textsc{IMDB} & 0.500 & 0.500 & 25,000
          & 15,000 & 10,000 & 25,000 \\
      \textsc{Kindle} & 0.917 & 0.083 & \phantom{0}3,821
          & \phantom{0}2,292 & \phantom{0}1,529 & 21,592 \\
      \textsc{HP} & 0.982 & 0.018 & \phantom{0}9,533
          & \phantom{0}5,720 & \phantom{0}3,813 & 18,401 \\
      \hline \end{tabular} } \vspace{-5ex}
  \end{center}
  \label{tab:datasets}
\end{table}

In our experiments, from each set of training data we randomly select
60\% of the documents for training purposes, leaving the remaining
40\% for the hyperparameter optimisation phase; these random splits
are stratified, meaning that the two resulting parts display the same
prevalence values as the set that originated them. In this phase (see
Section~\ref{sec:paropt}) we use $n=21$, $m=10$, and $q=500$, i.e., we
generate $m=10$ random samples of $q=500$ documents each, for each of
the $n=21$ prevalence values of the $\oplus$ class in
$\{0.00, 0.05, ..., 0.95, 1.00\}$.

In order to evaluate a quantifier over a wide spectrum of test
prevalence values, we use essentially the same process that we have
discussed in Section~\ref{sec:paropt} for hyperparameter optimisation;
that is, along with~\cite{Esuli:2018rm,Forman:2008kx}, we repeatedly
and randomly undersample one or both classes in the test set $U$ in
order to obtain testing samples with specified class prevalence
values. Here we generate $m=100$ random testing samples of $q=500$
documents each, for each of the $n=21$ prevalence values of the
$\oplus$ class in $\{0.00, 0.05, ..., 0.95, 1.00\}$.


\subsection{Evaluation measures}
\label{sec:measures}

\noindent As the measures of quantification error we use
\emph{Absolute Error} ($\aae$) and \emph{Relative Absolute Error}
($\rae$), defined as
\begin{align}
  \label{eq:aeandrae}
  \aae(p,\hat{p})  =\frac{1}{|\mathcal{Y}|}\sum_{y\in 
  \mathcal{Y}}|\hat{p}_{y}-p_{y}| 
  \hspace{3em}
  \rae(p,\hat{p})  =\frac{1}{|\mathcal{Y}|}\sum_{y\in 
  \mathcal{Y}}\displaystyle\frac{|\hat{p}_{y}-p_{y}|}{p_{y}} 
\end{align}
\noindent where $\mathcal{Y}$ is the set of classes of interest
($\mathcal{Y}=\{\oplus,\ominus\}$ in our case) \blue{and the sample
$\sigma$ is omitted for notational brevity}.  Note that $\rae$ is
undefined when at least one of the classes $y\in \mathcal{Y}$ is such
that its prevalence in $U$ is $0$. To solve this problem, in computing
$\rae$ we smooth both all $p_{y}$'s and $\hat{p}_{y}$'s via additive
smoothing, i.e., we take
$\underline{p}_{y}=\frac{\epsilon+p_{y}}{\sum_{y\in
\mathcal{Y}}(\epsilon+p_{y})}$,
where $\underline{p}_{y}$ denotes the smoothed version of $p_{y}$ and
the denominator is just a normalising factor (same for the
$\hat{\underline{p}}(y)$'s); following~\cite{Forman:2008kx}, we use
the quantity $\epsilon=\frac{1}{2 |U|}$ as the smoothing factor. We
then use the smoothed versions of $p_{y}$ and $\hat{p}_{y}$ in place
of their original non-smoothed versions in Equation~\ref{eq:aeandrae};
as a result, $\rae$ is always defined.

The reason why we use $\aae$ and $\rae$ is that from a theoretical
standpoint they are, as it has recently been
argued~\cite{Sebastiani:2020qf}, the most satisfactory evaluation
measures for quantification.


\subsection{Data processing}
We preprocess our documents by using the stop word remover and default
tokeniser available within the \texttt{scikit-learn}
framework\footnote{\url{http://scikit-learn.org/}}. In all three
datasets we remove all terms occurring less than 5 times in the
training set and all punctuation marks, and lowercase the text.
As the weighting criterion we use a version of the well-known
$\mathrm{tfidf}$ method, i.e.,
\begin{equation}
  \mathrm{tfidf}(f,\mathbf{x})=\log(\#(f,\mathbf{x})+1)\times 
  \log \frac{|L|}{|\mathbf{x}'\in L : \#(f,\mathbf{x}')>0|}
  \label{eq:tfidf}
\end{equation}
\noindent where $\#(f,\mathbf{x})$ is the raw number of occurrences of
feature $f$ in document $\mathbf{x}$; weights are then normalised via
cosine normalisation.

Among the learners we use for classification (see below), the only one
that does not rely on a $\mathrm{tfidf}$-based representation is
CNN. This learner simply converts all documents into lists of unique
numeric IDs, indexing the terms in the vocabulary. We pad the
documents to the first 300 words.


\subsection{The quantifiers}
\label{sec:quantifiers}

\noindent We here describe all the quantification systems we have used
in this work.

\subsubsection{CC and its variants.}
\label{sec:cc}

\noindent In our experiments we generate versions of CC, ACC, PCC, and
PACC, using five different learners, i.e., support vector machines
(SVM), logistic regression (LR), random forests (RF), multinomial
naive Bayes (MNB), and convolutional neural networks (CNN). For the
first four learners we rely on the implementations available from
\texttt{scikit-learn}, while the CNN deep neural network is something
we have implemented ourselves using the \texttt{pytorch}
framework.\footnote{\texttt{https://pytorch.org/}} The setups that we
use for these learners are the following:

\begin{itemize}
\item SVM: We use soft-margin SVMs with linear kernel and L2
  regularisation, and we explicitly optimise the $C$ parameter (in the
  range $C \in \{10^{i}\}$ with $i\in \{-4, -3,\ldots , 4, 5\}$) that
  determines the tradeoff between the margin and the training error
  (default: $C=1$).  We also optimise the $J_{\oplus}$ and
  $J_{\ominus}$ ``rebalancing'' parameters, which determine whether to
  impose that misclassifying a $\oplus$ document has a different cost
  than misclassifying a $\ominus$ document (in this case one sets
  $J_{\oplus}=\frac{p_\ominus(L)}{p_\oplus(L)}$ and $J_{\ominus}=1$),
  or not (in this case one sets $J_{\oplus}=J_{\ominus}=1$, which is
  the default configuration)~\cite{Morik99a}.
  
\item LR: As in SVM, we use L2 regularisation, and we explicitly
  optimise the rebalancing parameters and the regularisation
  coefficient $C$ (default values are as in SVM).
\item RF: 
  we optimise the number of estimators in the range
  $\{10, 50, 100, 250, 500\}$, the max depth in
  $\{5, 15, 30, \max\},$\footnote{When the depth is set to ``max''
  then nodes are expanded until all leaves belong to the same class.},
  and the splitting function in $\{\mathrm{Gini},\mathrm{Entropy}\}$
  (default: (100, max, Gini)).
\item MNB: We use Laplace smoothing, and we optimise the additive
  factor $\alpha$ in the range $\{0.00,0.05,\ldots,0.95,1.00\}$
  (default: $\alpha=1$).
\item CNN: we use a single convolutional layer with $\gamma$ output
  channels for three window lengths of 3, 5, and 7 words. Each
  convolution is followed by a ReLU activation function and a
  max-pooling operation. All convolved outputs are then concatenated
  and processed by an affine transformation and a sigmoid activation
  that
  converts the outputs into posterior probabilities. We use the Adam
  optimiser (with learning rate $1E^{-3}$ and all other parameters at
  their default values) to minimise the balanced binary cross-entropy
  loss,
  set the batch size to 100, and train the net for 500 epochs, but we
  apply an early stop after 20 consecutive training epochs showing no
  improvement in terms of $\mathrm{F}_1$ for the minority class on the
  validation set.  We explore the dimensionality of the embedding
  space in the range $\{100,300\}$ (default: 100), the number of
  output channels $\gamma$ in $\{256, 512\}$ (default: 512), whether
  to apply dropout to the last layer (with a drop probability of 0.5)
  or not (default: ``yes''), and whether to apply weight decay (with a
  factor of $1E^{-4}$) or not (default: ``no'').
\end{itemize}

\noindent Since we perform hyperparameter optimisation via grid
search, the number of validations (i.e., combinations of
hyperparameters) that we perform amounts to 20 for SVMs, 20 for LR, 40
for RF, 21 for MNB, and 16 for CNN.

In the following, by the notation $M^{m}_{l}$ we will indicate
quantification method $M$ using learner $l$ whose parameters have been
optimised using measure $m$ (where $M^{\mathrm{\O}}_{l}$ indicates
that no optimisation at all has been carried out).  We will test, on
all three datasets, all combinations in which $M$ ranges on \{CC, ACC,
PCC, PACC\}, $l$ ranges on \{SVM, LR, RF, MNB, CNN\}, and $m$ ranges
on \{$\mathrm{A}, \mathrm{F_{1}}, \mathrm{AE}\}$, where $\mathrm{A}$
denotes vanilla accuracy, $\mathrm{F_{1}}$ is the well-known harmonic
mean of precision and recall, and AE is absolute error.  We stick to
the tradition of computing $\mathrm{F_{1}}$ with respect to the
minority class, which always turns out to be $\ominus$ in all three
datasets (this means that, e.g., the true positives of the contingency
table are the documents that the classifier assigns to $\ominus$ and
that indeed belong to $\ominus$).

Note that PCC requires the classifier to return posterior
probabilities.  Since SVMs does not produce posterior probabilities,
for $\mathrm{PCC}_{\mathrm{SVM}}$ and $\mathrm{PACC}_{\mathrm{SVM}}$
we calibrate the confidence scores that SVMs return by using Platt's
method \cite{Platt:2000fk}.


\subsubsection{Advanced quantification methods.}
\label{sec:advanced}

\noindent As the advanced methods that we test against CC and its
variants, we use a number of more sophisticated systems that have been
top-performers in the recent quantification literature.

\begin{itemize}

\item We use the Saerens-Latinne-Decaestecker method
  \cite{Esuli:2020le,Saerens:2002uq} (SLD), which consists of training
  a probabilistic classifier and then exploiting the EM algorithm to
  iteratively shift the estimation of $p_{y}(U)$
  from the one that maximises the likelihood on the training set to
  the one that maximises it on the test data. As the underlying
  learner for SLD we use LR, 
  since (as MNB) it returns posterior probabilities (which SLD needs),
  since these probabilities tend to be (differently from those
  returned by MNB) well-calibrated, and since LR is well-known to
  perform much better than MNB.

\item We use methods SVM(KLD), SVM(NKLD), SVM(Q), SVM(AE), SVM(RAE),
  from the ``structured output learning'' camp. Each of them is the
  result of instantiating the SVM$_{\mathrm{perf}}$ structured output
  learner~\cite{Joachims05} to optimise a different loss
  function. SVM(KLD)~\cite{Esuli:2015gh} minimises the
  Kullback-Leibler Divergence (KLD); SVM(NKLD)~\cite{Esuli:2014uq}
  minimises a version of KLD normalised via the logistic function;
  SVM(Q)~\cite{Barranquero:2015fr} minimises the harmonic mean of a
  classification-oriented loss (recall) and a quantification-oriented
  loss ($\rae$). We also add versions that minimise AE and RAE, since
  these latter are now, as indicated in Section~\ref{sec:measures},
  the evaluation measures for quantification considered most
  satisfactory, and the two used in this paper for evaluating the
  quantification accuracy of our systems. We optimise the $C$
  parameter of SVM$_{\mathrm{perf}}$ in the range $C \in \{10^{i}\}$,
  with $i\in \{-4, -3,\ldots 4, 5\}$. In this case we do not optimise
  the $J_{\oplus}$ and $J_{\ominus}$ ``rebalancing'' parameters since
  this option is not available in SVM$_{\mathrm{perf}}$.

\item We use the HDy method of \cite{Gonzalez-Castro:2013fk}. The
  method searches for the prevalence values that minimise the
  divergence (as measured via the Hellinger Distance) between two
  cumulative distributions of posterior probabilities returned by the
  classifier, one for the unlabelled examples and the other for a
  validation set.  The latter is a mixture of the distributions of
  posterior probabilities returned for the $\oplus$ and $\ominus$
  validation examples, respectively, where the parameters of the
  mixture are the sought class prevalence values.
  We use LR as the classifier for the same reasons as discussed for
  SLD.

\item We use the QuaNet system, a ``meta-''quantification method based
  on deep learning~\cite{Esuli:2018rm}.  QuaNet takes as input a list
  of document embeddings, together with and sorted by the
  classification scores returned by a classifier. A bidirectional LSTM
  processes this list and produces a quantification embedding that is
  then concatenated with a vector of predictions produced by an
  ensemble of simpler quantification methods (we here employ CC, ACC,
  PCC, PACC, and SLD).  The resulting vector passes through a set of
  fully connected layers (followed by ReLU activations and dropout)
  that return the estimated class prevalence values. We use CNN as the
  learner since, among the learners we use in this paper, it is the
  only one that returns both posterior probabilities and document
  embeddings (we use the last layer of the CNN as the document
  embedding). 
  We set the hidden size of the bidirectional LSTM to $128+128=256$
  and use two stacked layers. We also set the hidden sizes of the
  fully connected layers to 1024 and 512, and the dropout probability
  to 0.5.  We train the network for 500 epochs, but we apply early
  stopping with a patience of 10 consecutive validations without
  improvements in terms of mean square error (MSE). Each training
  epoch consists of 200 quantification predictions, each of which for
  a batch of 500 randomly drawn documents at a prevalence sampled from
  the uniform distribution.  In our case, validation epochs correspond
  to 21 quantification predictions for batches of 500 documents
  randomly sampled to have prevalence values
  $0.00, 0.05, \ldots, 0.95, 1.00$.
  We use Adam as the optimiser, with default parameters, to minimise
  MSE. In order to train QuaNet, we split (using a 40\%/40\%/20\%
  stratified split) the training set $L_{\mathrm{Tr}}$ in three sets
  $L_{\mathrm{Tr}}^{\mathrm{CTr}}$, for training the classifier;
  $L_{\mathrm{Tr}}^{\mathrm{QTr}}$, for training QuaNet; and
  $L_{\mathrm{Tr}}^{\mathrm{QVa}}$, for validating QuaNet. When
  optimising QuaNet we do not explore any additional hyperparameter
  apart from those for the CNN.

\item We also report results for \emph{Maximum Likelihood Probability
  Estimation} (MLPE), the trivial baseline for quantification which
  makes the IID assumption and thus simply assumes that
  $p_{\oplus}(U)$ is identical to the training prevalence
  $p_{\oplus}(L)$
  irrespectively of the set $U$.

\end{itemize}

\noindent Note that ACC, PACC, HDy, and QuaNet need to estimate their
own parameters on a validation set, which means that their performance
depends on exactly which documents this set consists of. In order to
mitigate the impact of this random choice, for these methods we run
each experiment 10 times, each time with a different random
choice. The results we report are the average scores across these 10
runs.




\subsection{Results}
\label{sec:results}

\begin{table}[t]\caption{Results showing how the quantification error of CC changes 
        according to the measure used in hyperparameter optimization; a 
        negative percentage indicates a reduction in error with respect to 
        using the method with default parameters. The background cell color
        indicates improvement (green) or deterioration (red), while its 
        tone intensity is proportional to the absolute magnitude. }\label{tab:CC} \resizebox{\textwidth}{!} {
    \begin{tabular}{|l||ll|ll||ll|ll||ll|ll|}
    \hline
    & \multicolumn{4}{c||}{\textsc{IMDB}} 
    & \multicolumn{4}{c||}{\textsc{Kindle}} 
    & \multicolumn{4}{c|}{\textsc{HP}} \\ 
    \hline
    & \multicolumn{2}{c}{AE} 
    & \multicolumn{2}{|c||}{RAE} 
    & \multicolumn{2}{c}{AE} 
    & \multicolumn{2}{|c||}{RAE} 
    & \multicolumn{2}{c}{AE} 
    & \multicolumn{2}{|c|}{RAE} \\
    \hline
    CC$^{\mathrm{\O}}_{\mathrm{SVM}}$ & 0.065 & & 6.029 & & 0.305 & & 15.928 & & 0.471 & & 24.058 & \\
	CC$^{\mathrm{A}}_{\mathrm{SVM}}$ & \cellcolor{green!4}0.059 & \cellcolor{green!4} (-9.6\%)& \cellcolor{green!5}5.408 & \cellcolor{green!5} (-10.3\%)& \cellcolor{green!9}0.245 & \cellcolor{green!9} (-19.8\%)& \cellcolor{green!8}13.220 & \cellcolor{green!8} (-17.0\%)& \cellcolor{green!7}0.401 & \cellcolor{green!7} (-14.9\%)& \cellcolor{green!7}20.645 & \cellcolor{green!7} (-14.2\%)\\
	CC$^{\mathrm{F_1}}_{\mathrm{SVM}}$ & \cellcolor{green!4}0.059 & \cellcolor{green!4} (-9.5\%)& \cellcolor{green!4}5.523 & \cellcolor{green!4} (-8.4\%)& \cellcolor{green!32}0.108 & \cellcolor{green!32} (-64.5\%)& \cellcolor{green!27}7.192 & \cellcolor{green!27} (-54.8\%)& \cellcolor{green!24}0.236 & \cellcolor{green!24} (-50.0\%)& \cellcolor{green!21}13.590 & \cellcolor{green!21} (-43.5\%)\\
	CC$^{\mathrm{AE}}_{\mathrm{SVM}}$ & \cellcolor{red!0}0.065 & \cellcolor{red!0} (+0.3\%)& \cellcolor{red!0}6.091 & \cellcolor{red!0} (+1.0\%)& \cellcolor{green!33}0.100 & \cellcolor{green!33} (-67.1\%)& \cellcolor{green!26}7.555 & \cellcolor{green!26} (-52.6\%)& \cellcolor{green!37}0.119 & \cellcolor{green!37} (-74.8\%)& \cellcolor{green!27}10.593 & \cellcolor{green!27} (-56.0\%)\\
	\hline

	CC$^{\mathrm{\O}}_{\mathrm{LR}}$ & 0.059 & & 5.477 & & 0.470 & & 23.990 & & 0.500 & & 25.508 & \\
	CC$^{\mathrm{A}}_{\mathrm{LR}}$ & \cellcolor{red!3}0.062 & \cellcolor{red!3} (+6.0\%)& \cellcolor{red!3}5.839 & \cellcolor{red!3} (+6.6\%)& \cellcolor{green!28}0.202 & \cellcolor{green!28} (-57.0\%)& \cellcolor{green!26}11.215 & \cellcolor{green!26} (-53.3\%)& \cellcolor{green!4}0.451 & \cellcolor{green!4} (-9.8\%)& \cellcolor{green!4}23.035 & \cellcolor{green!4} (-9.7\%)\\
	CC$^{\mathrm{F_1}}_{\mathrm{LR}}$ & \cellcolor{red!2}0.062 & \cellcolor{red!2} (+5.3\%)& \cellcolor{red!2}5.725 & \cellcolor{red!2} (+4.5\%)& \cellcolor{green!32}0.163 & \cellcolor{green!32} (-65.3\%)& \cellcolor{green!30}9.278 & \cellcolor{green!30} (-61.3\%)& \cellcolor{green!27}0.229 & \cellcolor{green!27} (-54.3\%)& \cellcolor{green!23}13.505 & \cellcolor{green!23} (-47.1\%)\\
	CC$^{\mathrm{AE}}_{\mathrm{LR}}$ & \cellcolor{red!3}0.062 & \cellcolor{red!3} (+6.1\%)& \cellcolor{red!2}5.745 & \cellcolor{red!2} (+4.9\%)& \cellcolor{green!40}0.094 & \cellcolor{green!40} (-80.0\%)& \cellcolor{green!35}7.087 & \cellcolor{green!35} (-70.5\%)& \cellcolor{green!38}0.110 & \cellcolor{green!38} (-78.0\%)& \cellcolor{green!29}10.304 & \cellcolor{green!29} (-59.6\%)\\
	\hline

	CC$^{\mathrm{\O}}_{\mathrm{RF}}$ & 0.155 & & 13.388 & & 0.448 & & 22.988 & & 0.493 & & 25.196 & \\
	CC$^{\mathrm{A}}_{\mathrm{RF}}$ & \cellcolor{green!24}0.080 & \cellcolor{green!24} (-48.1\%)& \cellcolor{green!22}7.446 & \cellcolor{green!22} (-44.4\%)& \cellcolor{red!1}0.463 & \cellcolor{red!1} (+3.5\%)& \cellcolor{red!1}23.744 & \cellcolor{red!1} (+3.3\%)& \cellcolor{red!0}0.500 & \cellcolor{red!0} (+1.3\%)& \cellcolor{red!0}25.482 & \cellcolor{red!0} (+1.1\%)\\
	CC$^{\mathrm{F_1}}_{\mathrm{RF}}$ & \cellcolor{green!24}0.079 & \cellcolor{green!24} (-49.1\%)& \cellcolor{green!22}7.396 & \cellcolor{green!22} (-44.8\%)& \cellcolor{red!0}0.451 & \cellcolor{red!0} (+0.7\%)& \cellcolor{red!0}23.142 & \cellcolor{red!0} (+0.7\%)& \cellcolor{red!0}0.499 & \cellcolor{red!0} (+1.2\%)& \cellcolor{red!0}25.469 & \cellcolor{red!0} (+1.1\%)\\
	CC$^{\mathrm{AE}}_{\mathrm{RF}}$ & \cellcolor{green!24}0.079 & \cellcolor{green!24} (-48.8\%)& \cellcolor{green!22}7.487 & \cellcolor{green!22} (-44.1\%)& \cellcolor{red!1}0.464 & \cellcolor{red!1} (+3.6\%)& \cellcolor{red!1}23.721 & \cellcolor{red!1} (+3.2\%)& \cellcolor{red!0}0.500 & \cellcolor{red!0} (+1.3\%)& \cellcolor{red!0}25.487 & \cellcolor{red!0} (+1.2\%)\\
	\hline

	CC$^{\mathrm{\O}}_{\mathrm{MNB}}$ & 0.096 & & 8.147 & & 0.500 & & 25.513 & & 0.500 & & 25.510 & \\
	CC$^{\mathrm{A}}_{\mathrm{MNB}}$ & \cellcolor{red!0}0.098 & \cellcolor{red!0} (+1.6\%)& \cellcolor{red!2}8.529 & \cellcolor{red!2} (+4.7\%)& \cellcolor{green!5}0.443 & \cellcolor{green!5} (-11.4\%)& \cellcolor{green!5}22.641 & \cellcolor{green!5} (-11.3\%)& \cellcolor{green!0}0.499 & \cellcolor{green!0} (-0.2\%)& \cellcolor{green!0}25.459 & \cellcolor{green!0} (-0.2\%)\\
	CC$^{\mathrm{F_1}}_{\mathrm{MNB}}$ & \cellcolor{red!0}0.097 & \cellcolor{red!0} (+0.8\%)& \cellcolor{red!1}8.311 & \cellcolor{red!1} (+2.0\%)& \cellcolor{green!5}0.444 & \cellcolor{green!5} (-11.3\%)& \cellcolor{green!5}22.731 & \cellcolor{green!5} (-10.9\%)& \cellcolor{green!0}0.499 & \cellcolor{green!0} (-0.2\%)& \cellcolor{green!0}25.470 & \cellcolor{green!0} (-0.2\%)\\
	CC$^{\mathrm{AE}}_{\mathrm{MNB}}$ & \cellcolor{red!0}0.097 & \cellcolor{red!0} (+0.9\%)& \cellcolor{red!1}8.431 & \cellcolor{red!1} (+3.5\%)& \cellcolor{green!5}0.443 & \cellcolor{green!5} (-11.4\%)& \cellcolor{green!5}22.701 & \cellcolor{green!5} (-11.0\%)& \cellcolor{green!0}0.499 & \cellcolor{green!0} (-0.2\%)& \cellcolor{green!0}25.464 & \cellcolor{green!0} (-0.2\%)\\
	\hline

	CC$^{\mathrm{\O}}_{\mathrm{CNN}}$ & 0.072 & & 6.683 & & 0.087 & & 8.138 & & 0.255 & & 17.042 & \\
	CC$^{\mathrm{A}}_{\mathrm{CNN}}$ & \cellcolor{red!1}0.073 & \cellcolor{red!1} (+2.0\%)& \cellcolor{green!0}6.620 & \cellcolor{green!0} (-1.0\%)& \cellcolor{red!11}0.107 & \cellcolor{red!11} (+23.8\%)& \cellcolor{red!3}8.680 & \cellcolor{red!3} (+6.7\%)& \cellcolor{green!18}0.159 & \cellcolor{green!18} (-37.5\%)& \cellcolor{green!8}14.255 & \cellcolor{green!8} (-16.4\%)\\
	CC$^{\mathrm{F_1}}_{\mathrm{CNN}}$ & \cellcolor{red!4}0.078 & \cellcolor{red!4} (+8.7\%)& \cellcolor{red!3}7.142 & \cellcolor{red!3} (+6.9\%)& \cellcolor{green!1}0.085 & \cellcolor{green!1} (-2.2\%)& \cellcolor{green!1}7.951 & \cellcolor{green!1} (-2.3\%)& \cellcolor{green!20}0.149 & \cellcolor{green!20} (-41.5\%)& \cellcolor{green!8}14.030 & \cellcolor{green!8} (-17.7\%)\\
	CC$^{\mathrm{AE}}_{\mathrm{CNN}}$ & \cellcolor{red!1}0.074 & \cellcolor{red!1} (+3.2\%)& \cellcolor{green!0}6.613 & \cellcolor{green!0} (-1.0\%)& \cellcolor{red!13}0.109 & \cellcolor{red!13} (+26.2\%)& \cellcolor{red!2}8.591 & \cellcolor{red!2} (+5.6\%)& \cellcolor{red!17}0.343 & \cellcolor{red!17} (+34.3\%)& \cellcolor{red!5}19.008 & \cellcolor{red!5} (+11.5\%)\\
	\hline

	\end{tabular}
}
\end{table}

\begin{table}[t]\caption{Same as Table~\ref{tab:CC}, but with ACC instead of CC.}\label{tab:ACC} \resizebox{\textwidth}{!} {
    \begin{tabular}{|l||ll|ll||ll|ll||ll|ll|}
    \hline
    & \multicolumn{4}{c||}{\textsc{IMDB}} 
    & \multicolumn{4}{c||}{\textsc{Kindle}} 
    & \multicolumn{4}{c|}{\textsc{HP}} \\ 
    \hline
    & \multicolumn{2}{c}{AE} 
    & \multicolumn{2}{|c||}{RAE} 
    & \multicolumn{2}{c}{AE} 
    & \multicolumn{2}{|c||}{RAE} 
    & \multicolumn{2}{c}{AE} 
    & \multicolumn{2}{|c|}{RAE} \\
    \hline
    ACC$^{\mathrm{\O}}_{\mathrm{SVM}}$ & 0.023 & & 1.084 & & 0.068 & & 2.958 & & 0.341 & & 17.350 & \\
	ACC$^{\mathrm{A}}_{\mathrm{SVM}}$ & \cellcolor{green!8}0.019 & \cellcolor{green!8} (-17.6\%)& \cellcolor{green!8}0.889 & \cellcolor{green!8} (-18.0\%)& \cellcolor{red!2}0.070 & \cellcolor{red!2} (+4.1\%)& \cellcolor{red!2}3.093 & \cellcolor{red!2} (+4.6\%)& \cellcolor{green!23}0.181 & \cellcolor{green!23} (-47.0\%)& \cellcolor{green!23}9.245 & \cellcolor{green!23} (-46.7\%)\\
	ACC$^{\mathrm{F_1}}_{\mathrm{SVM}}$ & \cellcolor{green!2}0.022 & \cellcolor{green!2} (-5.2\%)& \cellcolor{red!3}1.153 & \cellcolor{red!3} (+6.3\%)& \cellcolor{green!11}0.052 & \cellcolor{green!11} (-22.9\%)& \cellcolor{green!10}2.309 & \cellcolor{green!10} (-21.9\%)& \cellcolor{green!33}0.110 & \cellcolor{green!33} (-67.8\%)& \cellcolor{green!29}7.019 & \cellcolor{green!29} (-59.5\%)\\
	ACC$^{\mathrm{AE}}_{\mathrm{SVM}}$ & \cellcolor{green!5}0.020 & \cellcolor{green!5} (-11.4\%)& \cellcolor{green!6}0.933 & \cellcolor{green!6} (-13.9\%)& \cellcolor{red!0}0.069 & \cellcolor{red!0} (+1.6\%)& \cellcolor{red!3}3.193 & \cellcolor{red!3} (+7.9\%)& \cellcolor{green!34}0.108 & \cellcolor{green!34} (-68.4\%)& \cellcolor{green!29}7.225 & \cellcolor{green!29} (-58.4\%)\\
	\hline

	ACC$^{\mathrm{\O}}_{\mathrm{LR}}$ & 0.017 & & 0.569 & & 0.279 & & 9.997 & & 0.500 & & 25.508 & \\
	ACC$^{\mathrm{A}}_{\mathrm{LR}}$ & \cellcolor{red!10}0.020 & \cellcolor{red!10} (+21.2\%)& \cellcolor{red!31}0.933 & \cellcolor{red!31} (+63.9\%)& \cellcolor{green!39}0.060 & \cellcolor{green!39} (-78.6\%)& \cellcolor{green!36}2.628 & \cellcolor{green!36} (-73.7\%)& \cellcolor{green!31}0.185 & \cellcolor{green!31} (-62.9\%)& \cellcolor{green!31}9.629 & \cellcolor{green!31} (-62.3\%)\\
	ACC$^{\mathrm{F_1}}_{\mathrm{LR}}$ & \cellcolor{red!7}0.019 & \cellcolor{red!7} (+15.9\%)& \cellcolor{red!28}0.896 & \cellcolor{red!28} (+57.4\%)& \cellcolor{green!39}0.057 & \cellcolor{green!39} (-79.5\%)& \cellcolor{green!37}2.507 & \cellcolor{green!37} (-74.9\%)& \cellcolor{green!40}0.098 & \cellcolor{green!40} (-80.5\%)& \cellcolor{green!37}6.534 & \cellcolor{green!37} (-74.4\%)\\
	ACC$^{\mathrm{AE}}_{\mathrm{LR}}$ & \cellcolor{red!5}0.018 & \cellcolor{red!5} (+10.8\%)& \cellcolor{red!24}0.850 & \cellcolor{red!24} (+49.3\%)& \cellcolor{green!38}0.065 & \cellcolor{green!38} (-76.9\%)& \cellcolor{green!35}2.891 & \cellcolor{green!35} (-71.1\%)& \cellcolor{green!40}0.092 & \cellcolor{green!40} (-81.7\%)& \cellcolor{green!38}5.849 & \cellcolor{green!38} (-77.1\%)\\
	\hline

	ACC$^{\mathrm{\O}}_{\mathrm{RF}}$ & 0.034 & & 1.254 & & 0.136 & & 4.199 & & 0.439 & & 23.528 & \\
	ACC$^{\mathrm{A}}_{\mathrm{RF}}$ & \cellcolor{green!19}0.021 & \cellcolor{green!19} (-38.7\%)& \cellcolor{green!24}0.643 & \cellcolor{green!24} (-48.8\%)& \cellcolor{red!15}0.180 & \cellcolor{red!15} (+31.7\%)& \cellcolor{red!28}6.603 & \cellcolor{red!28} (+57.3\%)& \cellcolor{red!4}0.482 & \cellcolor{red!4} (+9.7\%)& \cellcolor{red!2}24.654 & \cellcolor{red!2} (+4.8\%)\\
	ACC$^{\mathrm{F_1}}_{\mathrm{RF}}$ & \cellcolor{green!21}0.019 & \cellcolor{green!21} (-42.7\%)& \cellcolor{green!29}0.526 & \cellcolor{green!29} (-58.1\%)& \cellcolor{red!6}0.155 & \cellcolor{red!6} (+13.4\%)& \cellcolor{red!0}4.282 & \cellcolor{red!0} (+2.0\%)& \cellcolor{red!2}0.460 & \cellcolor{red!2} (+4.7\%)& \cellcolor{red!1}24.205 & \cellcolor{red!1} (+2.9\%)\\
	ACC$^{\mathrm{AE}}_{\mathrm{RF}}$ & \cellcolor{green!21}0.019 & \cellcolor{green!21} (-43.0\%)& \cellcolor{green!27}0.554 & \cellcolor{green!27} (-55.8\%)& \cellcolor{red!22}0.197 & \cellcolor{red!22} (+44.2\%)& \cellcolor{red!22}6.057 & \cellcolor{red!22} (+44.3\%)& \cellcolor{red!6}0.499 & \cellcolor{red!6} (+13.5\%)& \cellcolor{red!4}25.436 & \cellcolor{red!4} (+8.1\%)\\
	\hline

	ACC$^{\mathrm{\O}}_{\mathrm{MNB}}$ & 0.049 & & 2.316 & & 0.473 & & 23.280 & & 0.500 & & 25.508 & \\
	ACC$^{\mathrm{A}}_{\mathrm{MNB}}$ & \cellcolor{red!2}0.051 & \cellcolor{red!2} (+4.4\%)& \cellcolor{red!3}2.479 & \cellcolor{red!3} (+7.0\%)& \cellcolor{green!29}0.189 & \cellcolor{green!29} (-59.9\%)& \cellcolor{green!30}9.065 & \cellcolor{green!30} (-61.1\%)& \cellcolor{green!6}0.435 & \cellcolor{green!6} (-13.1\%)& \cellcolor{green!6}22.170 & \cellcolor{green!6} (-13.1\%)\\
	ACC$^{\mathrm{F_1}}_{\mathrm{MNB}}$ & \cellcolor{red!0}0.049 & \cellcolor{red!0} (+0.5\%)& \cellcolor{red!1}2.404 & \cellcolor{red!1} (+3.8\%)& \cellcolor{green!29}0.197 & \cellcolor{green!29} (-58.3\%)& \cellcolor{green!30}9.285 & \cellcolor{green!30} (-60.1\%)& \cellcolor{green!7}0.428 & \cellcolor{green!7} (-14.5\%)& \cellcolor{green!6}22.025 & \cellcolor{green!6} (-13.7\%)\\
	ACC$^{\mathrm{AE}}_{\mathrm{MNB}}$ & \cellcolor{red!1}0.051 & \cellcolor{red!1} (+3.9\%)& \cellcolor{red!5}2.591 & \cellcolor{red!5} (+11.9\%)& \cellcolor{green!27}0.213 & \cellcolor{green!27} (-54.9\%)& \cellcolor{green!27}10.376 & \cellcolor{green!27} (-55.4\%)& \cellcolor{green!4}0.451 & \cellcolor{green!4} (-9.7\%)& \cellcolor{green!4}23.146 & \cellcolor{green!4} (-9.3\%)\\
	\hline

	ACC$^{\mathrm{\O}}_{\mathrm{CNN}}$ & 0.021 & & 1.082 & & 0.074 & & 1.596 & & 0.173 & & 10.642 & \\
	ACC$^{\mathrm{A}}_{\mathrm{CNN}}$ & \cellcolor{green!4}0.019 & \cellcolor{green!4} (-8.2\%)& \cellcolor{green!12}0.811 & \cellcolor{green!12} (-25.0\%)& \cellcolor{green!6}0.064 & \cellcolor{green!6} (-12.7\%)& \cellcolor{green!2}1.515 & \cellcolor{green!2} (-5.1\%)& \cellcolor{red!14}0.223 & \cellcolor{red!14} (+28.6\%)& \cellcolor{green!3}9.939 & \cellcolor{green!3} (-6.6\%)\\
	ACC$^{\mathrm{F_1}}_{\mathrm{CNN}}$ & \cellcolor{red!5}0.023 & \cellcolor{red!5} (+10.1\%)& \cellcolor{green!0}1.067 & \cellcolor{green!0} (-1.4\%)& \cellcolor{green!8}0.061 & \cellcolor{green!8} (-17.4\%)& \cellcolor{green!5}1.424 & \cellcolor{green!5} (-10.8\%)& \cellcolor{red!2}0.182 & \cellcolor{red!2} (+5.3\%)& \cellcolor{green!1}10.344 & \cellcolor{green!1} (-2.8\%)\\
	ACC$^{\mathrm{AE}}_{\mathrm{CNN}}$ & \cellcolor{red!4}0.023 & \cellcolor{red!4} (+9.1\%)& \cellcolor{green!0}1.072 & \cellcolor{green!0} (-0.9\%)& \cellcolor{green!3}0.068 & \cellcolor{green!3} (-7.8\%)& \cellcolor{green!6}1.399 & \cellcolor{green!6} (-12.4\%)& \cellcolor{red!0}0.174 & \cellcolor{red!0} (+0.7\%)& \cellcolor{red!0}10.810 & \cellcolor{red!0} (+1.6\%)\\
	\hline

	\end{tabular}
}
\end{table}

\begin{table}[t]\caption{Same as Table~\ref{tab:CC}, but with PCC instead of CC.}\label{tab:PCC} \resizebox{\textwidth}{!} {
    \begin{tabular}{|l||ll|ll||ll|ll||ll|ll|}
    \hline
    & \multicolumn{4}{c||}{\textsc{IMDB}} 
    & \multicolumn{4}{c||}{\textsc{Kindle}} 
    & \multicolumn{4}{c|}{\textsc{HP}} \\ 
    \hline
    & \multicolumn{2}{c}{AE} 
    & \multicolumn{2}{|c||}{RAE} 
    & \multicolumn{2}{c}{AE} 
    & \multicolumn{2}{|c||}{RAE} 
    & \multicolumn{2}{c}{AE} 
    & \multicolumn{2}{|c|}{RAE} \\
    \hline
    PCC$^{\mathrm{\O}}_{\mathrm{SVM}}$ & 0.101 & & 9.460 & & 0.255 & & 14.514 & & 0.375 & & 20.158 & \\
	PCC$^{\mathrm{A}}_{\mathrm{SVM}}$ & \cellcolor{green!0}0.100 & \cellcolor{green!0} (-0.4\%)& \cellcolor{red!0}9.517 & \cellcolor{red!0} (+0.6\%)& \cellcolor{red!5}0.283 & \cellcolor{red!5} (+10.9\%)& \cellcolor{red!5}16.174 & \cellcolor{red!5} (+11.4\%)& \cellcolor{red!1}0.385 & \cellcolor{red!1} (+2.6\%)& \cellcolor{red!1}20.653 & \cellcolor{red!1} (+2.5\%)\\
	PCC$^{\mathrm{F_1}}_{\mathrm{SVM}}$ & \cellcolor{red!0}0.101 & \cellcolor{red!0} (+0.0\%)& \cellcolor{green!0}9.425 & \cellcolor{green!0} (-0.4\%)& \cellcolor{green!0}0.251 & \cellcolor{green!0} (-1.8\%)& \cellcolor{green!0}14.239 & \cellcolor{green!0} (-1.9\%)& \cellcolor{red!1}0.385 & \cellcolor{red!1} (+2.7\%)& \cellcolor{red!1}20.594 & \cellcolor{red!1} (+2.2\%)\\
	PCC$^{\mathrm{AE}}_{\mathrm{SVM}}$ & \cellcolor{green!0}0.100 & \cellcolor{green!0} (-0.4\%)& \cellcolor{red!0}9.484 & \cellcolor{red!0} (+0.2\%)& \cellcolor{green!0}0.254 & \cellcolor{green!0} (-0.6\%)& \cellcolor{green!0}14.461 & \cellcolor{green!0} (-0.4\%)& \cellcolor{red!1}0.386 & \cellcolor{red!1} (+2.8\%)& \cellcolor{red!1}20.607 & \cellcolor{red!1} (+2.2\%)\\
	\hline

	PCC$^{\mathrm{\O}}_{\mathrm{LR}}$ & 0.122 & & 11.564 & & 0.356 & & 20.405 & & 0.464 & & 24.608 & \\
	PCC$^{\mathrm{A}}_{\mathrm{LR}}$ & \cellcolor{green!12}0.091 & \cellcolor{green!12} (-25.5\%)& \cellcolor{green!12}8.563 & \cellcolor{green!12} (-26.0\%)& \cellcolor{green!10}0.279 & \cellcolor{green!10} (-21.5\%)& \cellcolor{green!13}15.031 & \cellcolor{green!13} (-26.3\%)& \cellcolor{green!12}0.352 & \cellcolor{green!12} (-24.2\%)& \cellcolor{green!12}18.605 & \cellcolor{green!12} (-24.4\%)\\
	PCC$^{\mathrm{F_1}}_{\mathrm{LR}}$ & \cellcolor{green!12}0.092 & \cellcolor{green!12} (-25.0\%)& \cellcolor{green!12}8.606 & \cellcolor{green!12} (-25.6\%)& \cellcolor{green!25}0.172 & \cellcolor{green!25} (-51.6\%)& \cellcolor{green!22}11.222 & \cellcolor{green!22} (-45.0\%)& \cellcolor{green!27}0.212 & \cellcolor{green!27} (-54.2\%)& \cellcolor{green!17}16.117 & \cellcolor{green!17} (-34.5\%)\\
	PCC$^{\mathrm{AE}}_{\mathrm{LR}}$ & \cellcolor{green!17}0.079 & \cellcolor{green!17} (-35.3\%)& \cellcolor{green!18}7.348 & \cellcolor{green!18} (-36.5\%)& \cellcolor{green!28}0.154 & \cellcolor{green!28} (-56.6\%)& \cellcolor{green!17}13.066 & \cellcolor{green!17} (-36.0\%)& \cellcolor{green!27}0.211 & \cellcolor{green!27} (-54.6\%)& \cellcolor{green!10}19.597 & \cellcolor{green!10} (-20.4\%)\\
	\hline

	PCC$^{\mathrm{\O}}_{\mathrm{RF}}$ & 0.199 & & 18.865 & & 0.376 & & 21.592 & & 0.461 & & 24.267 & \\
	PCC$^{\mathrm{A}}_{\mathrm{RF}}$ & \cellcolor{green!0}0.198 & \cellcolor{green!0} (-0.7\%)& \cellcolor{green!0}18.753 & \cellcolor{green!0} (-0.6\%)& \cellcolor{green!0}0.368 & \cellcolor{green!0} (-2.0\%)& \cellcolor{green!0}21.209 & \cellcolor{green!0} (-1.8\%)& \cellcolor{red!2}0.482 & \cellcolor{red!2} (+4.7\%)& \cellcolor{red!2}25.349 & \cellcolor{red!2} (+4.5\%)\\
	PCC$^{\mathrm{F_1}}_{\mathrm{RF}}$ & \cellcolor{green!1}0.195 & \cellcolor{green!1} (-2.1\%)& \cellcolor{green!1}18.459 & \cellcolor{green!1} (-2.2\%)& \cellcolor{green!0}0.372 & \cellcolor{green!0} (-0.9\%)& \cellcolor{green!0}21.319 & \cellcolor{green!0} (-1.3\%)& \cellcolor{red!0}0.466 & \cellcolor{red!0} (+1.1\%)& \cellcolor{red!0}24.563 & \cellcolor{red!0} (+1.2\%)\\
	PCC$^{\mathrm{AE}}_{\mathrm{RF}}$ & \cellcolor{green!0}0.196 & \cellcolor{green!0} (-1.4\%)& \cellcolor{green!0}18.565 & \cellcolor{green!0} (-1.6\%)& \cellcolor{green!1}0.366 & \cellcolor{green!1} (-2.5\%)& \cellcolor{green!1}21.088 & \cellcolor{green!1} (-2.3\%)& \cellcolor{red!0}0.462 & \cellcolor{red!0} (+0.3\%)& \cellcolor{red!0}24.379 & \cellcolor{red!0} (+0.5\%)\\
	\hline

	PCC$^{\mathrm{\O}}_{\mathrm{MNB}}$ & 0.171 & & 15.928 & & 0.478 & & 24.702 & & 0.498 & & 25.453 & \\
	PCC$^{\mathrm{A}}_{\mathrm{MNB}}$ & \cellcolor{green!0}0.168 & \cellcolor{green!0} (-1.7\%)& \cellcolor{green!0}15.663 & \cellcolor{green!0} (-1.7\%)& \cellcolor{green!10}0.381 & \cellcolor{green!10} (-20.3\%)& \cellcolor{green!8}20.396 & \cellcolor{green!8} (-17.4\%)& \cellcolor{green!0}0.497 & \cellcolor{green!0} (-0.2\%)& \cellcolor{green!0}25.397 & \cellcolor{green!0} (-0.2\%)\\
	PCC$^{\mathrm{F_1}}_{\mathrm{MNB}}$ & \cellcolor{green!1}0.167 & \cellcolor{green!1} (-2.2\%)& \cellcolor{green!0}15.617 & \cellcolor{green!0} (-2.0\%)& \cellcolor{green!10}0.380 & \cellcolor{green!10} (-20.4\%)& \cellcolor{green!8}20.369 & \cellcolor{green!8} (-17.5\%)& \cellcolor{green!2}0.473 & \cellcolor{green!2} (-5.0\%)& \cellcolor{green!1}24.487 & \cellcolor{green!1} (-3.8\%)\\
	PCC$^{\mathrm{AE}}_{\mathrm{MNB}}$ & \cellcolor{green!3}0.160 & \cellcolor{green!3} (-6.4\%)& \cellcolor{green!3}14.907 & \cellcolor{green!3} (-6.4\%)& \cellcolor{green!10}0.380 & \cellcolor{green!10} (-20.4\%)& \cellcolor{green!8}20.396 & \cellcolor{green!8} (-17.4\%)& \cellcolor{green!2}0.473 & \cellcolor{green!2} (-5.0\%)& \cellcolor{green!1}24.479 & \cellcolor{green!1} (-3.8\%)\\
	\hline

	PCC$^{\mathrm{\O}}_{\mathrm{CNN}}$ & 0.110 & & 9.994 & & 0.111 & & 10.448 & & 0.257 & & 18.368 & \\
	PCC$^{\mathrm{A}}_{\mathrm{CNN}}$ & \cellcolor{green!2}0.105 & \cellcolor{green!2} (-4.8\%)& \cellcolor{green!0}9.893 & \cellcolor{green!0} (-1.0\%)& \cellcolor{red!19}0.154 & \cellcolor{red!19} (+39.2\%)& \cellcolor{red!1}10.775 & \cellcolor{red!1} (+3.1\%)& \cellcolor{red!25}0.389 & \cellcolor{red!25} (+51.6\%)& \cellcolor{red!7}21.093 & \cellcolor{red!7} (+14.8\%)\\
	PCC$^{\mathrm{F_1}}_{\mathrm{CNN}}$ & \cellcolor{green!5}0.099 & \cellcolor{green!5} (-10.3\%)& \cellcolor{green!3}9.377 & \cellcolor{green!3} (-6.2\%)& \cellcolor{red!0}0.111 & \cellcolor{red!0} (+0.3\%)& \cellcolor{green!4}9.474 & \cellcolor{green!4} (-9.3\%)& \cellcolor{green!1}0.251 & \cellcolor{green!1} (-2.2\%)& \cellcolor{green!3}17.005 & \cellcolor{green!3} (-7.4\%)\\
	PCC$^{\mathrm{AE}}_{\mathrm{CNN}}$ & \cellcolor{red!15}0.145 & \cellcolor{red!15} (+31.3\%)& \cellcolor{red!5}11.146 & \cellcolor{red!5} (+11.5\%)& \cellcolor{red!16}0.148 & \cellcolor{red!16} (+33.8\%)& \cellcolor{red!17}14.017 & \cellcolor{red!17} (+34.2\%)& \cellcolor{green!19}0.156 & \cellcolor{green!19} (-39.3\%)& \cellcolor{green!10}14.644 & \cellcolor{green!10} (-20.3\%)\\
	\hline

	\end{tabular}
}
\end{table}

\begin{table}[t]\caption{Same as Table~\ref{tab:CC}, but with PACC instead of CC.}\label{tab:PACC} \resizebox{\textwidth}{!} {
    \begin{tabular}{|l||ll|ll||ll|ll||ll|ll|}
    \hline
    & \multicolumn{4}{c||}{\textsc{IMDB}} 
    & \multicolumn{4}{c||}{\textsc{Kindle}} 
    & \multicolumn{4}{c|}{\textsc{HP}} \\ 
    \hline
    & \multicolumn{2}{c}{AE} 
    & \multicolumn{2}{|c||}{RAE} 
    & \multicolumn{2}{c}{AE} 
    & \multicolumn{2}{|c||}{RAE} 
    & \multicolumn{2}{c}{AE} 
    & \multicolumn{2}{|c|}{RAE} \\
    \hline
    PACC$^{\mathrm{\O}}_{\mathrm{SVM}}$ & 0.021 & & 1.166 & & 0.059 & & 2.464 & & 0.137 & & 8.368 & \\
	PACC$^{\mathrm{A}}_{\mathrm{SVM}}$ & \cellcolor{green!1}0.021 & \cellcolor{green!1} (-3.2\%)& \cellcolor{red!2}1.215 & \cellcolor{red!2} (+4.3\%)& \cellcolor{red!5}0.065 & \cellcolor{red!5} (+10.0\%)& \cellcolor{red!8}2.893 & \cellcolor{red!8} (+17.4\%)& \cellcolor{green!11}0.106 & \cellcolor{green!11} (-22.8\%)& \cellcolor{green!11}6.425 & \cellcolor{green!11} (-23.2\%)\\
	PACC$^{\mathrm{F_1}}_{\mathrm{SVM}}$ & \cellcolor{green!1}0.021 & \cellcolor{green!1} (-3.4\%)& \cellcolor{red!1}1.202 & \cellcolor{red!1} (+3.1\%)& \cellcolor{red!5}0.066 & \cellcolor{red!5} (+11.4\%)& \cellcolor{red!10}2.979 & \cellcolor{red!10} (+20.9\%)& \cellcolor{red!4}0.148 & \cellcolor{red!4} (+8.2\%)& \cellcolor{red!2}8.723 & \cellcolor{red!2} (+4.2\%)\\
	PACC$^{\mathrm{AE}}_{\mathrm{SVM}}$ & \cellcolor{red!2}0.022 & \cellcolor{red!2} (+5.1\%)& \cellcolor{red!8}1.363 & \cellcolor{red!8} (+17.0\%)& \cellcolor{green!0}0.059 & \cellcolor{green!0} (-1.4\%)& \cellcolor{green!2}2.333 & \cellcolor{green!2} (-5.3\%)& \cellcolor{green!8}0.114 & \cellcolor{green!8} (-16.6\%)& \cellcolor{green!5}7.497 & \cellcolor{green!5} (-10.4\%)\\
	\hline

	PACC$^{\mathrm{\O}}_{\mathrm{LR}}$ & 0.017 & & 0.846 & & 0.064 & & 2.456 & & 0.119 & & 9.639 & \\
	PACC$^{\mathrm{A}}_{\mathrm{LR}}$ & \cellcolor{red!11}0.021 & \cellcolor{red!11} (+22.0\%)& \cellcolor{red!14}1.087 & \cellcolor{red!14} (+28.4\%)& \cellcolor{green!8}0.053 & \cellcolor{green!8} (-16.7\%)& \cellcolor{green!5}2.177 & \cellcolor{green!5} (-11.4\%)& \cellcolor{red!11}0.147 & \cellcolor{red!11} (+23.1\%)& \cellcolor{green!6}8.316 & \cellcolor{green!6} (-13.7\%)\\
	PACC$^{\mathrm{F_1}}_{\mathrm{LR}}$ & \cellcolor{red!12}0.021 & \cellcolor{red!12} (+24.5\%)& \cellcolor{red!19}1.176 & \cellcolor{red!19} (+39.0\%)& \cellcolor{red!1}0.065 & \cellcolor{red!1} (+2.2\%)& \cellcolor{green!8}2.060 & \cellcolor{green!8} (-16.1\%)& \cellcolor{green!11}0.091 & \cellcolor{green!11} (-23.2\%)& \cellcolor{green!9}7.748 & \cellcolor{green!9} (-19.6\%)\\
	PACC$^{\mathrm{AE}}_{\mathrm{LR}}$ & \cellcolor{red!13}0.021 & \cellcolor{red!13} (+26.5\%)& \cellcolor{red!23}1.237 & \cellcolor{red!23} (+46.3\%)& \cellcolor{red!2}0.068 & \cellcolor{red!2} (+5.5\%)& \cellcolor{green!4}2.253 & \cellcolor{green!4} (-8.3\%)& \cellcolor{green!6}0.104 & \cellcolor{green!6} (-12.3\%)& \cellcolor{green!4}8.812 & \cellcolor{green!4} (-8.6\%)\\
	\hline

	PACC$^{\mathrm{\O}}_{\mathrm{RF}}$ & 0.030 & & 1.221 & & 0.074 & & 2.923 & & 0.168 & & 10.322 & \\
	PACC$^{\mathrm{A}}_{\mathrm{RF}}$ & \cellcolor{green!14}0.022 & \cellcolor{green!14} (-28.4\%)& \cellcolor{green!14}0.877 & \cellcolor{green!14} (-28.2\%)& \cellcolor{red!5}0.082 & \cellcolor{red!5} (+10.4\%)& \cellcolor{red!7}3.367 & \cellcolor{red!7} (+15.2\%)& \cellcolor{red!3}0.180 & \cellcolor{red!3} (+7.1\%)& \cellcolor{red!3}11.095 & \cellcolor{red!3} (+7.5\%)\\
	PACC$^{\mathrm{F_1}}_{\mathrm{RF}}$ & \cellcolor{green!14}0.021 & \cellcolor{green!14} (-29.8\%)& \cellcolor{green!11}0.952 & \cellcolor{green!11} (-22.0\%)& \cellcolor{red!3}0.079 & \cellcolor{red!3} (+6.9\%)& \cellcolor{red!6}3.331 & \cellcolor{red!6} (+13.9\%)& \cellcolor{green!2}0.160 & \cellcolor{green!2} (-5.1\%)& \cellcolor{red!0}10.350 & \cellcolor{red!0} (+0.3\%)\\
	PACC$^{\mathrm{AE}}_{\mathrm{RF}}$ & \cellcolor{green!16}0.020 & \cellcolor{green!16} (-33.2\%)& \cellcolor{green!12}0.914 & \cellcolor{green!12} (-25.1\%)& \cellcolor{red!4}0.081 & \cellcolor{red!4} (+8.9\%)& \cellcolor{red!6}3.286 & \cellcolor{red!6} (+12.4\%)& \cellcolor{green!8}0.140 & \cellcolor{green!8} (-17.1\%)& \cellcolor{green!1}10.067 & \cellcolor{green!1} (-2.5\%)\\
	\hline

	PACC$^{\mathrm{\O}}_{\mathrm{MNB}}$ & 0.055 & & 3.253 & & 0.180 & & 7.352 & & 0.195 & & 10.930 & \\
	PACC$^{\mathrm{A}}_{\mathrm{MNB}}$ & \cellcolor{red!2}0.058 & \cellcolor{red!2} (+4.8\%)& \cellcolor{red!2}3.412 & \cellcolor{red!2} (+4.9\%)& \cellcolor{green!13}0.130 & \cellcolor{green!13} (-27.7\%)& \cellcolor{green!8}6.058 & \cellcolor{green!8} (-17.6\%)& \cellcolor{red!35}0.335 & \cellcolor{red!35} (+71.6\%)& \cellcolor{red!31}17.883 & \cellcolor{red!31} (+63.6\%)\\
	PACC$^{\mathrm{F_1}}_{\mathrm{MNB}}$ & \cellcolor{red!4}0.060 & \cellcolor{red!4} (+8.1\%)& \cellcolor{red!3}3.487 & \cellcolor{red!3} (+7.2\%)& \cellcolor{green!16}0.122 & \cellcolor{green!16} (-32.2\%)& \cellcolor{green!12}5.570 & \cellcolor{green!12} (-24.2\%)& \cellcolor{red!42}0.363 & \cellcolor{red!42} (+86.0\%)& \cellcolor{red!32}18.138 & \cellcolor{red!32} (+65.9\%)\\
	PACC$^{\mathrm{AE}}_{\mathrm{MNB}}$ & \cellcolor{red!7}0.063 & \cellcolor{red!7} (+14.9\%)& \cellcolor{red!8}3.815 & \cellcolor{red!8} (+17.3\%)& \cellcolor{green!9}0.144 & \cellcolor{green!9} (-19.6\%)& \cellcolor{green!4}6.626 & \cellcolor{green!4} (-9.9\%)& \cellcolor{red!13}0.248 & \cellcolor{red!13} (+27.2\%)& \cellcolor{red!14}13.999 & \cellcolor{red!14} (+28.1\%)\\
	\hline

	PACC$^{\mathrm{\O}}_{\mathrm{CNN}}$ & 0.022 & & 1.205 & & 0.064 & & 1.414 & & 0.181 & & 9.808 & \\
	PACC$^{\mathrm{A}}_{\mathrm{CNN}}$ & \cellcolor{green!5}0.019 & \cellcolor{green!5} (-11.1\%)& \cellcolor{green!9}0.970 & \cellcolor{green!9} (-19.5\%)& \cellcolor{red!11}0.079 & \cellcolor{red!11} (+23.0\%)& \cellcolor{red!8}1.664 & \cellcolor{red!8} (+17.7\%)& \cellcolor{green!5}0.161 & \cellcolor{green!5} (-11.3\%)& \cellcolor{green!2}9.293 & \cellcolor{green!2} (-5.3\%)\\
	PACC$^{\mathrm{F_1}}_{\mathrm{CNN}}$ & \cellcolor{green!7}0.019 & \cellcolor{green!7} (-14.4\%)& \cellcolor{green!11}0.928 & \cellcolor{green!11} (-23.0\%)& \cellcolor{red!6}0.073 & \cellcolor{red!6} (+13.0\%)& \cellcolor{red!1}1.464 & \cellcolor{red!1} (+3.5\%)& \cellcolor{green!3}0.169 & \cellcolor{green!3} (-6.5\%)& \cellcolor{green!3}9.034 & \cellcolor{green!3} (-7.9\%)\\
	PACC$^{\mathrm{AE}}_{\mathrm{CNN}}$ & \cellcolor{green!8}0.018 & \cellcolor{green!8} (-17.3\%)& \cellcolor{green!15}0.830 & \cellcolor{green!15} (-31.2\%)& \cellcolor{red!3}0.069 & \cellcolor{red!3} (+6.9\%)& \cellcolor{green!1}1.367 & \cellcolor{green!1} (-3.3\%)& \cellcolor{green!4}0.165 & \cellcolor{green!4} (-9.1\%)& \cellcolor{green!4}8.829 & \cellcolor{green!4} (-10.0\%)\\
	\hline

	\end{tabular}
}
\end{table}

\noindent Tables~\ref{tab:CC}, \ref{tab:ACC}, \ref{tab:PCC}, and
\ref{tab:PACC} report the results obtained for CC, ACC, PCC, and
PACC. At a first glance, the results do not seem to give any clearcut
indication on how the CC variants should be optimised.  However, a
closer look reveals a number of patterns. One of these is that
SVM and LR (the two best-performing classifiers overall) tend to
benefit from optimised hyperparameters, and tend to do so to a greater
extent when the loss used in the optimisation is
quantification-oriented. Somehow surprisingly, not all methods improve
after model selection in every case.  However, there tends to be such
an improvement especially for ACC and PACC.  A likely reason for this
is the possible existence of a complex tradeoff between obtaining a
more accurate classifier and obtaining more reliable estimates for the
TPR and FPR quantities.

Regarding the different datasets, it seems that there is no clear
improvement from performing model selection when the training set is
balanced (see \textsc{IMDB}), neither by using a
classification-oriented measure
nor by using a quantification-oriented one.  A possible reason is
that any classifier (with or without hyperparameter optimisation)
becomes a reasonable quantifier if it learns to pay equal importance
to positive and negative examples, i.e., if the errors it produces are
unbiased towards either $\oplus$ or $\ominus$.  In this respect, RF
and MNB prove strongly biased towards the majority class, and only
when corrected via an adjustment (ACC or PACC) they deliver results
comparable to those obtained for other learners.

CNN works well on average almost in all cases, and seems to be the
least sensitive learner to model selection.

In order to better understand whether or not, on average and across
different situations, CC and its variants benefit from performing
model selection using a quantification-oriented loss, we have
submitted our results to a statistical significance test.
Table~\ref{tab:stats} shows the outcome of a two-sided t-test on
\emph{related} sets of scores, across datasets and learners, from
which we can compare pairs of model selection methods.  The test
reveals that optimising AE works better than optimising A or than
using default settings ($\O$).  The test does not clearly say whether
optimising AE or $\mathrm{F}_1$ is better, but it suggests that PACC
(the strongest CC variant) works better when optimised for AE than
when optimised for $\mathrm{F}_1$.

\begin{table}[t]
  \caption{
  Two-sided t-test results on \emph{related} samples of error scores across datasets and learners.
  For a pair of optimization measures X vs.\ Y, symbol $\gg$ (resp. $>$) indicates that method X performs
  better (i.e., yields lower error) than Y, and that the difference in performance, as averaged across 
  pairs of experiments on all datasets 
  and learners, is statistically significant at a confidence score of $\alpha=0.001$ (resp. $\alpha=0.05$).
  Symbols $\ll$ and $<$ have a similar meaning but indicate that X performs worse (i.e., yields higher error) than Y.
  Symbol $\sim$ instead indicates that the differences in performance between X and Y are not
  statistically significantly different, i.e., that $p$-value $\geq 0.05$. 
  }
  \label{tab:stats}
  \center

{\scriptsize   \begin{tabular}{|rcl||c|c||c|c||c|c||c|c|}
    \hline
    \multicolumn{3}{|c||}{\mbox{}}
    & \multicolumn{2}{c||}{\textsc{CC}} 
    & \multicolumn{2}{c||}{\textsc{ACC}} 
    & \multicolumn{2}{c|}{\textsc{PCC}} 
    & \multicolumn{2}{c|}{\textsc{PACC}}\\ 
    \cline{4-11}
    \multicolumn{3}{|c||}{\mbox{}}
    & \multicolumn{1}{c|}{AE} 
    & \multicolumn{1}{c||}{RAE} 
    & \multicolumn{1}{c|}{AE} 
    & \multicolumn{1}{c||}{RAE} 
    & \multicolumn{1}{c|}{AE} 
    & \multicolumn{1}{c|}{RAE}
    & \multicolumn{1}{c|}{AE} 
    & \multicolumn{1}{c|}{RAE} \\
    \hline

    $\mathrm{AE}$ & vs & $\mathrm{F_1}$ & $\gg$ & $\sim$ & $\ll$ & $\ll$ & $\gg$ & $\ll$ & $\gg$ & $\gg$\\
$\mathrm{AE}$ & vs & $\mathrm{A}$ & $\gg$ & $\gg$ & $\gg$ & $>$ & $\gg$ & $\gg$ & $\gg$ & $\gg$\\
$\mathrm{AE}$ & vs & $\mathrm{\O}$ & $\gg$ & $\gg$ & $\gg$ & $\gg$ & $\gg$ & $\gg$ & $\gg$ & $\sim$\\
$\mathrm{F_1}$ & vs & $\mathrm{A}$ & $\gg$ & $\gg$ & $\gg$ & $\gg$ & $\gg$ & $\gg$ & $\sim$ & $\sim$\\
$\mathrm{F_1}$ & vs & $\mathrm{\O}$ & $\gg$ & $\gg$ & $\gg$ & $\gg$ & $\gg$ & $\gg$ & $\ll$ & $\ll$\\
$\mathrm{A}$ & vs & $\mathrm{\O}$ & $\gg$ & $\gg$ & $\gg$ & $\gg$ & $\gg$ & $\gg$ & $\ll$ & $\ll$\\

    \hline
  \end{tabular}
 } 
\end{table}

Finally, Table \ref{tab:overview} compares the CC variants against
more recent state-of-the-art quantification systems. Columns AE and
RAE indicate the error of each method for each dataset. Columns
r$_{\mathrm{AE}}$ and r$_{\mathrm{RAE}}$ show the rank positions for
each pair (dataset, error) and, in parentheses, the rank position each
method would have obtained in case the CC variants had not been
optimised.

Interestingly, although some advanced quantification methods
(specifically: SLD and HDy) stand as the top performers, many among
the (supposedly more sophisticated) quantification methods fail to
improve over CC's performance. At a glance, most quantification
methods tend to obtain lower ranks when compared with properly
optimised CC variants. Remarkable examples of rank variation include
CC and ACC with SVM and LR: when evaluated on \textsc{Kindle} and
\textsc{HP}, they climb several positions (up to 25), often
entering the group of the 10 top-performing methods. In the most
extreme case, $\mathrm{ACC}_{\mathrm{LR}}^{\mathrm{AE}}$ moves from
position 28 (out of 29) to position 3 once properly optimised for
quantification.

\begin{table}[t]
  \caption{Results showing how CC and its variants, once optimised
  using a quantification-oriented measure, compare with more modern
  quantification methods. \textbf{Boldface} indicates the best method.
  \blue{For columns AE and RAE, the best/worst results are highlighted
  in bright green/red; the colour for the other scores is a linearly
  interpolation between these two extremes. For columns
  r$_{\mathrm{AE}}$ and r$_{\mathrm{RAE}}$, green/red is used to
  denote methods which have obtained higher/lower rank positions once
  the CC variants have been optimised for AE, with respect to the case
  in which they have not been optimised at all.}  All scores are
  different, in a statistically significant sense, from the best one
  according to a paired sample, two-tailed t-test at a confidence
  level of $0.001$.  }
  \label{tab:overview}
  \center

  \resizebox{\textwidth}{!} { 
  \begin{tabular}{|l|l||r|r||r|r||r|r||rr|rr||rr|rr||rr|rr|}
    \hline
    \multicolumn{2}{|c||}{\mbox{}}
    & \multicolumn{2}{c||}{\textsc{IMDB}} 
    & \multicolumn{2}{c||}{\textsc{Kindle}} 
    & \multicolumn{2}{c||}{\textsc{HP}} 
    & \multicolumn{4}{c||}{\textsc{IMDB}} 
    & \multicolumn{4}{c||}{\textsc{Kindle}} 
    & \multicolumn{4}{c|}{\textsc{HP}} \\ 
    \cline{3-20}
    \multicolumn{2}{|c||}{\mbox{}}
    & \multicolumn{1}{c|}{AE} 
    & \multicolumn{1}{c||}{RAE} 
    & \multicolumn{1}{c|}{AE} 
    & \multicolumn{1}{c||}{RAE} 
    & \multicolumn{1}{c|}{AE} 
    & \multicolumn{1}{c||}{RAE} 
    & \multicolumn{2}{c|}{r$_{\mathrm{AE}}$} 
    & \multicolumn{2}{c||}{r$_{\mathrm{RAE}}$} 
    & \multicolumn{2}{c|}{r$_{\mathrm{AE}}$} 
    & \multicolumn{2}{c||}{r$_{\mathrm{RAE}}$} 
    & \multicolumn{2}{c|}{r$_{\mathrm{AE}}$} 
    & \multicolumn{2}{c|}{r$_{\mathrm{RAE}}$} \\
    \hline
    \multirow{20}{*}{\begin{sideways}CC and its variants\end{sideways}}
 & CC$^{\mathrm{AE}}_{\mathrm{SVM}}$ & 0.065\cellcolor{green!29}& 6.091\cellcolor{green!26}& 0.100\cellcolor{green!37}& 7.555\cellcolor{green!23}& 0.119\cellcolor{green!33}& 10.593\cellcolor{green!8}& 20 \cellcolor{green!0}& (20) \cellcolor{green!0}& 20 \cellcolor{green!0}& (20) \cellcolor{green!0}& 13 \cellcolor{green!13}& (21) \cellcolor{green!13}& 15 \cellcolor{green!10}& (21) \cellcolor{green!10}& 8 \cellcolor{green!24}& (22) \cellcolor{green!24}& 11 \cellcolor{green!15}& (20) \cellcolor{green!15}\\
	 & ACC$^{\mathrm{AE}}_{\mathrm{SVM}}$ & 0.020\cellcolor{green!47}& 0.933\cellcolor{green!47}& 0.069\cellcolor{green!45}& 3.193\cellcolor{green!41}& 0.108\cellcolor{green!35}& 7.225\cellcolor{green!22}& 7 \cellcolor{green!1}& (8) \cellcolor{green!1}& 7 \cellcolor{red!1}& (6) \cellcolor{red!1}& 8 \cellcolor{red!3}& (6) \cellcolor{red!3}& 9 \cellcolor{green!0}& (9) \cellcolor{green!0}& 5 \cellcolor{green!18}& (16) \cellcolor{green!18}& 4 \cellcolor{green!18}& (15) \cellcolor{green!18}\\
	 & PCC$^{\mathrm{AE}}_{\mathrm{SVM}}$ & 0.100\cellcolor{green!15}& 9.484\cellcolor{green!12}& 0.254\cellcolor{green!0}& 14.461\cellcolor{red!5}& 0.386\cellcolor{red!25}& 20.607\cellcolor{red!30}& 25 \cellcolor{red!3}& (23) \cellcolor{red!3}& 25 \cellcolor{red!3}& (23) \cellcolor{red!3}& 24 \cellcolor{red!8}& (19) \cellcolor{red!8}& 24 \cellcolor{red!6}& (20) \cellcolor{red!6}& 21 \cellcolor{red!6}& (17) \cellcolor{red!6}& 22 \cellcolor{red!6}& (18) \cellcolor{red!6}\\
	 & PACC$^{\mathrm{AE}}_{\mathrm{SVM}}$ & 0.022\cellcolor{green!46}& 1.363\cellcolor{green!45}& 0.059\cellcolor{green!47}& 2.333\cellcolor{green!44}& 0.114\cellcolor{green!34}& 7.497\cellcolor{green!21}& 9 \cellcolor{red!5}& (6) \cellcolor{red!5}& 11 \cellcolor{red!6}& (7) \cellcolor{red!6}& 3 \cellcolor{green!0}& (3) \cellcolor{green!0}& 7 \cellcolor{green!0}& (7) \cellcolor{green!0}& 7 \cellcolor{red!3}& (5) \cellcolor{red!3}& 5 \cellcolor{red!3}& (3) \cellcolor{red!3}\\
	\cline{2-20}
 & CC$^{\mathrm{AE}}_{\mathrm{LR}}$ & 0.062\cellcolor{green!30}& 5.745\cellcolor{green!27}& 0.094\cellcolor{green!39}& 7.087\cellcolor{green!24}& 0.110\cellcolor{green!35}& 10.304\cellcolor{green!10}& 14 \cellcolor{green!0}& (14) \cellcolor{green!0}& 15 \cellcolor{red!1}& (14) \cellcolor{red!1}& 12 \cellcolor{green!24}& (26) \cellcolor{green!24}& 14 \cellcolor{green!20}& (26) \cellcolor{green!20}& 6 \cellcolor{green!39}& (29) \cellcolor{green!39}& 10 \cellcolor{green!31}& (28) \cellcolor{green!31}\\
	 & ACC$^{\mathrm{AE}}_{\mathrm{LR}}$ & 0.018\cellcolor{green!48}& 0.850\cellcolor{green!47}& 0.065\cellcolor{green!46}& 2.891\cellcolor{green!42}& 0.092\cellcolor{green!39}& 5.849\cellcolor{green!27}& 4 \cellcolor{red!3}& (2) \cellcolor{red!3}& 5 \cellcolor{red!3}& (3) \cellcolor{red!3}& 4 \cellcolor{green!27}& (20) \cellcolor{green!27}& 8 \cellcolor{green!15}& (17) \cellcolor{green!15}& 3 \cellcolor{green!43}& (28) \cellcolor{green!43}& 3 \cellcolor{green!41}& (27) \cellcolor{green!41}\\
	 & PCC$^{\mathrm{AE}}_{\mathrm{LR}}$ & 0.079\cellcolor{green!23}& 7.348\cellcolor{green!21}& 0.154\cellcolor{green!24}& 13.066\cellcolor{green!0}& 0.211\cellcolor{green!13}& 19.597\cellcolor{red!26}& 22 \cellcolor{green!5}& (25) \cellcolor{green!5}& 22 \cellcolor{green!5}& (25) \cellcolor{green!5}& 19 \cellcolor{green!5}& (22) \cellcolor{green!5}& 22 \cellcolor{green!0}& (22) \cellcolor{green!0}& 16 \cellcolor{green!8}& (21) \cellcolor{green!8}& 20 \cellcolor{green!3}& (22) \cellcolor{green!3}\\
	 & PACC$^{\mathrm{AE}}_{\mathrm{LR}}$ & 0.021\cellcolor{green!47}& 1.237\cellcolor{green!45}& 0.068\cellcolor{green!45}& 2.253\cellcolor{green!44}& 0.104\cellcolor{green!36}& 8.812\cellcolor{green!15}& 8 \cellcolor{red!8}& (3) \cellcolor{red!8}& 10 \cellcolor{red!10}& (4) \cellcolor{red!10}& 5 \cellcolor{red!1}& (4) \cellcolor{red!1}& 6 \cellcolor{green!0}& (6) \cellcolor{green!0}& 4 \cellcolor{red!1}& (3) \cellcolor{red!1}& 6 \cellcolor{red!1}& (5) \cellcolor{red!1}\\
	\cline{2-20}
 & CC$^{\mathrm{AE}}_{\mathrm{RF}}$ & 0.079\cellcolor{green!23}& 7.487\cellcolor{green!20}& 0.464\cellcolor{red!50}& 23.721\cellcolor{red!43}& 0.500\cellcolor{red!50}& 25.487\cellcolor{red!50}& 23 \cellcolor{green!5}& (26) \cellcolor{green!5}& 23 \cellcolor{green!5}& (26) \cellcolor{green!5}& 29 \cellcolor{red!6}& (25) \cellcolor{red!6}& 28 \cellcolor{red!6}& (24) \cellcolor{red!6}& 29 \cellcolor{red!8}& (24) \cellcolor{red!8}& 29 \cellcolor{red!10}& (23) \cellcolor{red!10}\\
	 & ACC$^{\mathrm{AE}}_{\mathrm{RF}}$ & 0.019\cellcolor{green!47}& 0.554\cellcolor{green!48}& 0.197\cellcolor{green!14}& 6.057\cellcolor{green!29}& 0.499\cellcolor{red!49}& 25.436\cellcolor{red!49}& 5 \cellcolor{green!10}& (11) \cellcolor{green!10}& 3 \cellcolor{green!13}& (11) \cellcolor{green!13}& 21 \cellcolor{red!12}& (14) \cellcolor{red!12}& 11 \cellcolor{red!1}& (10) \cellcolor{red!1}& 27 \cellcolor{red!13}& (19) \cellcolor{red!13}& 26 \cellcolor{red!12}& (19) \cellcolor{red!12}\\
	 & PCC$^{\mathrm{AE}}_{\mathrm{RF}}$ & 0.196\cellcolor{red!23}& 18.565\cellcolor{red!24}& 0.366\cellcolor{red!26}& 21.088\cellcolor{red!32}& 0.462\cellcolor{red!41}& 24.379\cellcolor{red!45}& 28 \cellcolor{green!0}& (28) \cellcolor{green!0}& 28 \cellcolor{green!0}& (28) \cellcolor{green!0}& 25 \cellcolor{red!3}& (23) \cellcolor{red!3}& 26 \cellcolor{red!5}& (23) \cellcolor{red!5}& 24 \cellcolor{red!6}& (20) \cellcolor{red!6}& 24 \cellcolor{red!5}& (21) \cellcolor{red!5}\\
	 & PACC$^{\mathrm{AE}}_{\mathrm{RF}}$ & 0.020\cellcolor{green!47}& 0.914\cellcolor{green!47}& 0.081\cellcolor{green!42}& 3.286\cellcolor{green!40}& 0.140\cellcolor{green!28}& 10.067\cellcolor{green!10}& 6 \cellcolor{green!6}& (10) \cellcolor{green!6}& 6 \cellcolor{green!6}& (10) \cellcolor{green!6}& 10 \cellcolor{red!1}& (9) \cellcolor{red!1}& 10 \cellcolor{red!3}& (8) \cellcolor{red!3}& 10 \cellcolor{red!6}& (6) \cellcolor{red!6}& 9 \cellcolor{red!3}& (7) \cellcolor{red!3}\\
	\cline{2-20}
 & CC$^{\mathrm{AE}}_{\mathrm{MNB}}$ & 0.097\cellcolor{green!16}& 8.431\cellcolor{green!16}& 0.443\cellcolor{red!44}& 22.701\cellcolor{red!39}& 0.499\cellcolor{red!49}& 25.464\cellcolor{red!49}& 24 \cellcolor{red!3}& (22) \cellcolor{red!3}& 24 \cellcolor{red!3}& (22) \cellcolor{red!3}& 28 \cellcolor{green!1}& (29) \cellcolor{green!1}& 27 \cellcolor{green!3}& (29) \cellcolor{green!3}& 28 \cellcolor{red!3}& (26) \cellcolor{red!3}& 28 \cellcolor{green!1}& (29) \cellcolor{green!1}\\
	 & ACC$^{\mathrm{AE}}_{\mathrm{MNB}}$ & 0.051\cellcolor{green!35}& 2.591\cellcolor{green!40}& 0.213\cellcolor{green!10}& 10.376\cellcolor{green!11}& 0.451\cellcolor{red!39}& 23.146\cellcolor{red!40}& 12 \cellcolor{green!0}& (12) \cellcolor{green!0}& 12 \cellcolor{green!0}& (12) \cellcolor{green!0}& 23 \cellcolor{green!6}& (27) \cellcolor{green!6}& 20 \cellcolor{green!8}& (25) \cellcolor{green!8}& 23 \cellcolor{green!6}& (27) \cellcolor{green!6}& 23 \cellcolor{green!5}& (26) \cellcolor{green!5}\\
	 & PCC$^{\mathrm{AE}}_{\mathrm{MNB}}$ & 0.160\cellcolor{red!8}& 14.907\cellcolor{red!9}& 0.380\cellcolor{red!29}& 20.396\cellcolor{red!29}& 0.473\cellcolor{red!44}& 24.479\cellcolor{red!46}& 27 \cellcolor{green!0}& (27) \cellcolor{green!0}& 27 \cellcolor{green!0}& (27) \cellcolor{green!0}& 26 \cellcolor{green!3}& (28) \cellcolor{green!3}& 25 \cellcolor{green!3}& (27) \cellcolor{green!3}& 25 \cellcolor{green!0}& (25) \cellcolor{green!0}& 25 \cellcolor{green!0}& (25) \cellcolor{green!0}\\
	 & PACC$^{\mathrm{AE}}_{\mathrm{MNB}}$ & 0.063\cellcolor{green!30}& 3.815\cellcolor{green!35}& 0.144\cellcolor{green!26}& 6.626\cellcolor{green!26}& 0.248\cellcolor{green!4}& 13.999\cellcolor{red!4}& 16 \cellcolor{red!5}& (13) \cellcolor{red!5}& 13 \cellcolor{green!0}& (13) \cellcolor{green!0}& 16 \cellcolor{green!1}& (17) \cellcolor{green!1}& 12 \cellcolor{green!0}& (12) \cellcolor{green!0}& 19 \cellcolor{red!15}& (10) \cellcolor{red!15}& 17 \cellcolor{red!13}& (9) \cellcolor{red!13}\\
	\cline{2-20}
 & CC$^{\mathrm{AE}}_{\mathrm{CNN}}$ & 0.074\cellcolor{green!25}& 6.613\cellcolor{green!24}& 0.109\cellcolor{green!35}& 8.591\cellcolor{green!18}& 0.343\cellcolor{red!15}& 19.008\cellcolor{red!24}& 21 \cellcolor{green!0}& (21) \cellcolor{green!0}& 21 \cellcolor{green!0}& (21) \cellcolor{green!0}& 14 \cellcolor{red!5}& (11) \cellcolor{red!5}& 17 \cellcolor{red!5}& (14) \cellcolor{red!5}& 20 \cellcolor{red!10}& (14) \cellcolor{red!10}& 19 \cellcolor{red!8}& (14) \cellcolor{red!8}\\
	 & ACC$^{\mathrm{AE}}_{\mathrm{CNN}}$ & 0.023\cellcolor{green!46}& 1.072\cellcolor{green!46}& 0.068\cellcolor{green!45}& 1.399\cellcolor{green!48}& 0.174\cellcolor{green!21}& 10.810\cellcolor{green!8}& 10 \cellcolor{red!8}& (5) \cellcolor{red!8}& 8 \cellcolor{red!5}& (5) \cellcolor{red!5}& 6 \cellcolor{green!3}& (8) \cellcolor{green!3}& 3 \cellcolor{green!0}& (3) \cellcolor{green!0}& 13 \cellcolor{red!10}& (7) \cellcolor{red!10}& 12 \cellcolor{red!6}& (8) \cellcolor{red!6}\\
	 & PCC$^{\mathrm{AE}}_{\mathrm{CNN}}$ & 0.145\cellcolor{red!2}& 11.146\cellcolor{green!5}& 0.148\cellcolor{green!25}& 14.017\cellcolor{red!3}& 0.156\cellcolor{green!25}& 14.644\cellcolor{red!7}& 26 \cellcolor{red!3}& (24) \cellcolor{red!3}& 26 \cellcolor{red!3}& (24) \cellcolor{red!3}& 17 \cellcolor{red!8}& (12) \cellcolor{red!8}& 23 \cellcolor{red!8}& (18) \cellcolor{red!8}& 11 \cellcolor{green!6}& (15) \cellcolor{green!6}& 18 \cellcolor{red!3}& (16) \cellcolor{red!3}\\
	 & PACC$^{\mathrm{AE}}_{\mathrm{CNN}}$ & 0.018\cellcolor{green!48}& 0.830\cellcolor{green!47}& 0.069\cellcolor{green!45}& 1.367\cellcolor{green!48}& 0.165\cellcolor{green!23}& 8.829\cellcolor{green!15}& 2 \cellcolor{green!8}& (7) \cellcolor{green!8}& 4 \cellcolor{green!8}& (9) \cellcolor{green!8}& 7 \cellcolor{red!3}& (5) \cellcolor{red!3}& 2 \cellcolor{green!0}& (2) \cellcolor{green!0}& 12 \cellcolor{red!6}& (8) \cellcolor{red!6}& 7 \cellcolor{red!1}& (6) \cellcolor{red!1}\\
	\cline{2-20}
\hline\hline\multirow{9}{*}{\begin{sideways}Baselines\end{sideways}} & SLD$^{\mathrm{AE}}_{\mathrm{LR}}$ & \textbf{0.014}\cellcolor{green!50}& \textbf{0.216}\cellcolor{green!50}& \textbf{0.048}\cellcolor{green!50}& 1.606\cellcolor{green!47}& \textbf{0.042}\cellcolor{green!50}& \textbf{0.195}\cellcolor{green!50}& \textbf{1}\cellcolor{green!0}& \textbf{(1)}\cellcolor{green!0}& \textbf{1}\cellcolor{green!0}& \textbf{(1)}\cellcolor{green!0}& \textbf{1}\cellcolor{green!0}& \textbf{(1)}\cellcolor{green!0}& 4 \cellcolor{green!0}& (4) \cellcolor{green!0}& \textbf{1}\cellcolor{green!0}& \textbf{(1)}\cellcolor{green!0}& \textbf{1}\cellcolor{green!0}& \textbf{(1)}\cellcolor{green!0}\\
	 & SVM(KLD)$^{\mathrm{AE}}_{\mathrm{}}$ & 0.064\cellcolor{green!29}& 5.936\cellcolor{green!26}& 0.122\cellcolor{green!32}& 7.866\cellcolor{green!21}& 0.185\cellcolor{green!18}& 12.185\cellcolor{green!2}& 18 \cellcolor{green!0}& (18) \cellcolor{green!0}& 18 \cellcolor{green!0}& (18) \cellcolor{green!0}& 15 \cellcolor{red!3}& (13) \cellcolor{red!3}& 16 \cellcolor{red!5}& (13) \cellcolor{red!5}& 14 \cellcolor{red!8}& (9) \cellcolor{red!8}& 14 \cellcolor{red!5}& (11) \cellcolor{red!5}\\
	 & SVM(NKLD)$^{\mathrm{AE}}_{\mathrm{}}$ & 0.065\cellcolor{green!29}& 5.927\cellcolor{green!26}& 0.085\cellcolor{green!41}& 6.693\cellcolor{green!26}& 0.121\cellcolor{green!32}& 9.566\cellcolor{green!12}& 19 \cellcolor{green!0}& (19) \cellcolor{green!0}& 16 \cellcolor{green!0}& (16) \cellcolor{green!0}& 11 \cellcolor{red!1}& (10) \cellcolor{red!1}& 13 \cellcolor{red!3}& (11) \cellcolor{red!3}& 9 \cellcolor{red!8}& (4) \cellcolor{red!8}& 8 \cellcolor{red!6}& (4) \cellcolor{red!6}\\
	 & SVM(Q)$^{\mathrm{AE}}_{\mathrm{}}$ & 0.064\cellcolor{green!30}& 5.928\cellcolor{green!26}& 0.208\cellcolor{green!11}& 11.384\cellcolor{green!7}& 0.386\cellcolor{red!25}& 19.956\cellcolor{red!28}& 17 \cellcolor{green!0}& (17) \cellcolor{green!0}& 17 \cellcolor{green!0}& (17) \cellcolor{green!0}& 22 \cellcolor{red!6}& (18) \cellcolor{red!6}& 21 \cellcolor{red!3}& (19) \cellcolor{red!3}& 22 \cellcolor{red!6}& (18) \cellcolor{red!6}& 21 \cellcolor{red!6}& (17) \cellcolor{red!6}\\
	 & SVM(AE)$^{\mathrm{AE}}_{\mathrm{}}$ & 0.060\cellcolor{green!31}& 5.572\cellcolor{green!28}& 0.159\cellcolor{green!23}& 9.705\cellcolor{green!14}& 0.219\cellcolor{green!11}& 13.090\cellcolor{red!0}& 13 \cellcolor{green!3}& (15) \cellcolor{green!3}& 14 \cellcolor{green!1}& (15) \cellcolor{green!1}& 20 \cellcolor{red!6}& (16) \cellcolor{red!6}& 19 \cellcolor{red!5}& (16) \cellcolor{red!5}& 17 \cellcolor{red!8}& (12) \cellcolor{red!8}& 15 \cellcolor{red!5}& (12) \cellcolor{red!5}\\
	 & SVM(RAE)$^{\mathrm{RAE}}_{\mathrm{}}$ & 0.063\cellcolor{green!30}& 5.957\cellcolor{green!26}& 0.152\cellcolor{green!24}& 9.242\cellcolor{green!16}& 0.239\cellcolor{green!6}& 13.575\cellcolor{red!2}& 15 \cellcolor{green!1}& (16) \cellcolor{green!1}& 19 \cellcolor{green!0}& (19) \cellcolor{green!0}& 18 \cellcolor{red!5}& (15) \cellcolor{red!5}& 18 \cellcolor{red!5}& (15) \cellcolor{red!5}& 18 \cellcolor{red!8}& (13) \cellcolor{red!8}& 16 \cellcolor{red!5}& (13) \cellcolor{red!5}\\
	 & HDy$^{\mathrm{AE}}_{\mathrm{LR}}$ & 0.018\cellcolor{green!48}& 0.420\cellcolor{green!49}& 0.055\cellcolor{green!48}& \textbf{1.027}\cellcolor{green!50}& 0.058\cellcolor{green!46}& 2.970\cellcolor{green!39}& 3 \cellcolor{green!1}& (4) \cellcolor{green!1}& 2 \cellcolor{green!0}& (2) \cellcolor{green!0}& 2 \cellcolor{green!0}& (2) \cellcolor{green!0}& \textbf{1}\cellcolor{green!0}& \textbf{(1)}\cellcolor{green!0}& 2 \cellcolor{green!0}& (2) \cellcolor{green!0}& 2 \cellcolor{green!0}& (2) \cellcolor{green!0}\\
	 & QuaNet$^{\mathrm{AE}}_{\mathrm{CNN}}$ & 0.027\cellcolor{green!44}& 1.175\cellcolor{green!46}& 0.070\cellcolor{green!44}& 2.119\cellcolor{green!45}& 0.210\cellcolor{green!13}& 11.433\cellcolor{green!5}& 11 \cellcolor{red!3}& (9) \cellcolor{red!3}& 9 \cellcolor{red!1}& (8) \cellcolor{red!1}& 9 \cellcolor{red!3}& (7) \cellcolor{red!3}& 5 \cellcolor{green!0}& (5) \cellcolor{green!0}& 15 \cellcolor{red!6}& (11) \cellcolor{red!6}& 13 \cellcolor{red!5}& (10) \cellcolor{red!5}\\
	\cline{2-20}
 & MLPE$^{\mathrm{\O}}_{\mathrm{\O}}$ & 0.262\cellcolor{red!50}& 24.874\cellcolor{red!50}& 0.429\cellcolor{red!41}& 25.266\cellcolor{red!50}& 0.484\cellcolor{red!46}& 25.447\cellcolor{red!49}& 29 \cellcolor{green!0}& (29) \cellcolor{green!0}& 29 \cellcolor{green!0}& (29) \cellcolor{green!0}& 27 \cellcolor{red!5}& (24) \cellcolor{red!5}& 29 \cellcolor{red!1}& (28) \cellcolor{red!1}& 26 \cellcolor{red!5}& (23) \cellcolor{red!5}& 27 \cellcolor{red!5}& (24) \cellcolor{red!5}\\
	
    \hline
  \end{tabular}
   }

\end{table}


\section{Conclusions}
\label{sec:conclusions}

\noindent One of the takeaway messages from the present work is that,
when using CC and/or its variants as baselines in their research on
learning to quantify, researchers should properly optimise these
baselines (i.e., use a truly quantification-oriented protocol, which
includes the use of a quantification-oriented loss, in hyperparameter
optimisation), lest these baselines become strawmen. The extensive
empirical evaluation we have carried out shows that, in general, the
performance of CC and its variants improves when the underlying
learner has been optimised with a quantification-oriented loss (AE).
The results of our experiments are less clear about whether optimising
AE or $\mathrm{F}_1$ (which, despite being a
\emph{classification}-oriented loss, is one that rewards classifiers
that balance FPs and FNs) is better, although they indicate that
optimising AE is preferable for PACC, the strongest among the variants
of CC.


\section*{Acknowledgments}
\label{sec:Acknowledgments}

\noindent 
The present work has been supported
by the \textsf{SoBigData++} project, funded by the European Commission
(Grant 871042) under the H2020 Programme INFRAIA-2019-1, and by the
\textsf{AI4Media} project, funded by the European Commission (Grant
951911) under the H2020 Programme ICT-48-2020. The authors' opinions
do not necessarily reflect those of the European Commission.




\end{document}